\pdfoutput=1
\documentclass[11pt]{article}

\usepackage[]{acl}

\usepackage{times}
\usepackage{latexsym}
\usepackage{comment}
\usepackage{ragged2e}
\usepackage{makecell}
\usepackage[T1]{fontenc}
\usepackage[utf8]{inputenc}
\usepackage{pifont}
\newcommand{\cmark}{\ding{51}}
\newcommand{\xmark}{\ding{55}}
\usepackage{microtype}

\usepackage{inconsolata}

\usepackage{inconsolata}
\usepackage{amssymb}
\usepackage{multirow}
\usepackage{tabularx}
\usepackage{caption}
\usepackage{graphicx}
\usepackage{subcaption}
\usepackage{booktabs}
\usepackage{algorithm}
\usepackage[noend]{algorithmic}
\usepackage{bm}
\title{PAT-Questions: A Self-Updating Benchmark for Present-Anchored Temporal Question-Answering}

\author{Jannat Ara Meem, Muhammad Shihab Rashid, Yue Dong, Vagelis Hristidis \\
    University of California, Riverside \\
    \texttt{\{jmeem001, mrash013, yue.dong\}@ucr.edu, vagelis@cs.ucr.edu}}
\begin{document}
\maketitle
\begin{abstract}
Existing work on Temporal Question Answering (TQA) has predominantly focused on questions anchored to specific timestamps or events (e.g. `Who was the US president in 1970?'). 
Little work has studied questions whose temporal context is relative to the present time (e.g. `Who was the previous US president?').
We refer to this problem as Present-Anchored Temporal QA (PATQA). 
PATQA poses unique challenges: (1) large language models (LLMs) may have outdated knowledge, (2) complex temporal relationships (e.g. `before', `previous') are hard to reason, (3) multi-hop reasoning may be required, and (4) the gold answers of benchmarks must be continuously updated. 
To address these challenges, we introduce the PAT-Questions benchmark, which includes single and multi-hop temporal questions. The answers in PAT-Questions can be automatically refreshed by re-running SPARQL queries on a knowledge graph, if available.
%capable of automatics answers' updates to stay aligned with the latest knowledge. Constructed using templates derived from time-dependent Wikidata facts, PAT-Questions ensures data quality and relevance by associating each question with a SPARQL query for up-to-date answer retrieval. 
We evaluate several state-of-the-art LLMs and a SOTA temporal reasoning model (TEMPREASON-T5) on PAT-Questions through direct prompting and retrieval-augmented generation (RAG).
% using Google Custom Search. 
The results highlight the limitations of existing solutions in PATQA and motivate the need for new methods to improve PATQA reasoning capabilities.
\end{abstract}

\section{Introduction}
% \begin{figure*}[t]
%     \centering
%     \includegraphics[width=\textwidth]{Figures/IntroFigure3.pdf}
%     \caption{Illustration of the limitations of the LLMs in answering the present-anchored temporal questions. Most of the LLMs respond with an out-of-date answer and some lack multi-hop temporal reasoning capabilities.}
%     \label{fig:enter-label}
% \end{figure*}

\begin{figure*}[t]
    \centering
    \begin{subfigure}[b]{0.6\linewidth}
      \centering
      \includegraphics[width=0.9\linewidth]{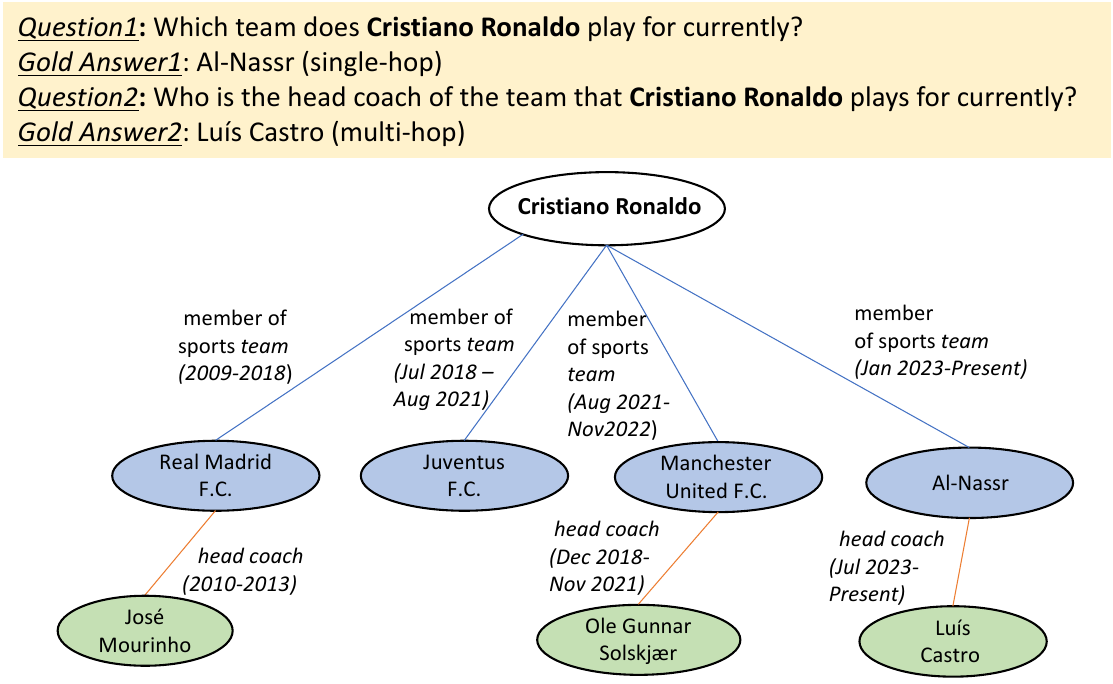}%test2_all.pdf}
      \caption{A subgraph from Wikidata around \textbf{Subject: Cristiano Ronaldo}}
      \label{fig:intro_wiki_graph}
    \end{subfigure}%
    ~
    \begin{subfigure}[b]{0.4\linewidth}
      \centering
      \includegraphics[width=0.9\linewidth]{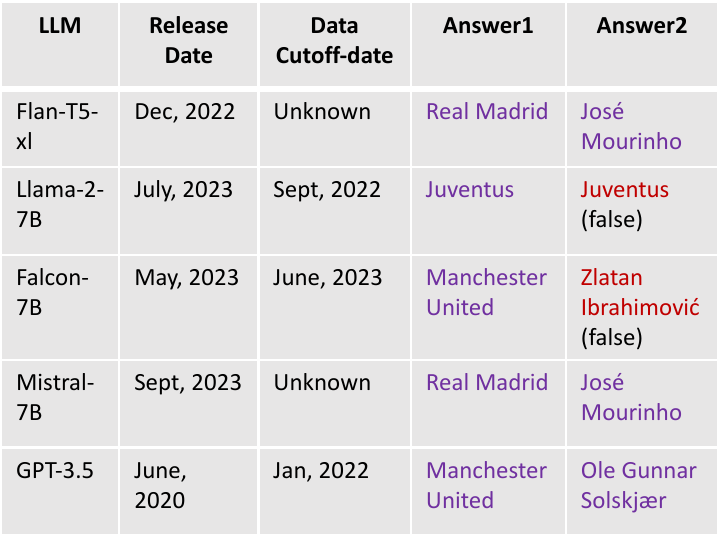}
      ~
      \caption{LLM responses for the two questions}
      \label{fig:intro_llm_answers}
    \end{subfigure}%
    \caption{Illustration of the limitations of the LLMs in answering the present-anchored temporal questions. The LLMs respond with an out-of-date answer (purple) due to knowledge outdating or a false information (red) due to lacking multi-hop PAT reasoning abilities.}
    \label{fig:intro}
\end{figure*}
\begin{comment}
In recent years, large language models (LLMs) have achieved significant progress in open-book question answering (OBQA) \citep{ye2023multi, zhao2023verify}, knowledge-base question answering (KBQA) \citep{tan2023can}, multi-hop reasoning \citep{kojima2022large, wang2023self} and temporal question answering (TSQA) \citep{dhingra2022time, jia2018tempquestions, tan2023towards} tasks.
\end{comment}
Large language models (LLMs) have demonstrated impressive performance across a wide spectrum of question-answering (QA) domains, thanks to an abundant amount of data spanning different QA tasks such as open-book question answering (OBQA) \citep{ye2023multi, zhao2023verify}, knowledge-base question answering (KBQA) \citep{tan2023can}, and multi-hop reasoning tasks~\citep{kojima2022large, wang2023self}. Their ability to tackle temporal question answering (TQA) has also seen considerable advancements, as evidenced by recent literature \citep{dhingra2022time, jia2018tempquestions, tan2023towards}.

However,  many past studies on Temporal Question Answering (TQA) focus on questions anchored by specific timestamps or events, such as `Who was the president of the US in 1985/during World War II?'.  In real life, we argue that many of the questions LLMs face are present-anchored without a specific timestamp, representing a crucial yet underexplored category of TQA.  We refer to it as Present-Anchored Temporal QA (PATQA), where the time condition is relative to the present time, for example, `Which team does Cristiano Ronaldo play for currently?'.

PATQA poses challenges due to several factors: (1) LLMs' knowledge becomes \textbf{outdated} due to periodic training \citep{he2022rethinking, zhang2021situatedqa, liska2022streamingqa}. Efforts to mitigate this through retrieval-augmented generation (RAG), providing current documents as context, are also often ineffective \citep{lewis2020retrieval, kasai2022realtime, vu2023freshllms}, as verified by our own experiments with New Bing\footnote{\url{https://www.bing.com/chat}} (using GPT-4, shown in Section \ref{sec:resuts}).
% LLMs are trained periodically, so their knowledge is outdated \citep{he2022rethinking, zhang2021situatedqa, liska2022streamingqa, kasai2022realtime}. Some studies have attempted to address this limitation using retrieval-augmented generation (RAG) \citep{lewis2020retrieval, vu2023freshllms} where up-to-date documents retrieved are provided as context to the LLMs. However, the effectiveness of document retrieval in alignment with temporal signals remains inadequate.
(2) PATQA can contain \textbf{complex temporal relationships} (e.g. before, last, previous) that are challenging. For example, `Which team did Cristiano Ronaldo play for before the current team?' requires a sequential understanding of temporal expressions `current' and `before'. (3) PATQA  may require \textbf{multi-hop}  reasoning that involves temporal reasoning in subsequent hops. For example, tracing Cristiano Ronaldo’s current team to its current head coach in `Who is the head coach of the team that Cristiano Ronaldo plays for currently?' (Cristiano Ronaldo $\rightarrow $ team $\rightarrow $ head coach), requires sequential temporal reasoning followed by multi-hop information integration  (head coaches change with time too). 
\textit{(4)} Creating and \textbf{maintaining PATQA benchmarks} is expensive because the gold answers to the questions keep changing and manual updates are not sustainable and scalable.
Figure \ref{fig:intro} illustrates examples that, due to these challenges, current LLMs perform poorly on data in our created PATQA dataset.

We introduce a novel benchmark, referred to as \textit{PAT-Questions}\footnote{\textbf{P}resent-\textbf{A}nchored \textbf{T}emporal \textbf{Questions}}, comprising 6172 present time-sensitive factual question-answer pairs that possess the four features we have mentioned above. These challenges require both single and multi-hop temporal reasoning over complex temporal relations to answer correctly. A unique property of PAT-Questions is its capability to automatically update answers over time, resulting in distinct instances for different timestamps. We construct PAT-Questions by leveraging templates derived from time-dependent facts sourced from the Wikidata knowledge base \citep{vrandevcic2014wikidata}. This allows us to ground our questions on Wikidata facts, thereby ensuring data quality over time by associating a SPARQL query with each question to accurately retrieve answers from the most up-to-date Wikidata.

As far as we know, there are only two datasets which contain present-anchored temporal QA examples, but without  complex temporal relations like `before', `previous', and have very few multi-hop temporal questions  \citep{kasai2022realtime, vu2023freshllms}. Further, \textit{these datasets do not offer a way to automatically update the answers over time}, which limits their applicability to future PATQA algorithms.

We benchmark several state-of-the-art (SOTA) LLMs on PAT-Questions, both directly prompting the LLMs with the questions, and in a RAG setting. To retieve documents in RAG, we use Google Custom Search (GCS), following \citet{kasai2022realtime}'s work, to retrieve relevant documents from up-to-date Wikipedia and Wikidata first, and provide the documents as context along with the initial prompt to the LLMs. We also evaluate the performance of a SOTA temporal reasoning system \citep{tan2023towards}, which fine-tunes the T5-SFT model \citep{raffel2020exploring}. In their setting, external context in the form of natural language text is provided. In contrast, we consider an open retrieval setting \citep{nguyen2016ms, rashid2024normy} to retrieve the most relevant context for each question and provide that as context to the LLMs. Our empirical results highlight that the SOTA models significantly
% We found that the PAT-Questions 
struggle on PAT-Questions, especially on multi-hop ones, with EM accuracy ranging from 1.5\% to 15.5\%. 
% \shihab{how can PATQ have low accuracy, do you mean LLMs?} \meem{Does it look better now?}
%The results highlight the limitation of the LLMs on PATQA and our goal is to elicit further research efforts on PATQA that can improve the LLMs' ability to answer present-anchored temporal questions.\\
\begin{table*}[t]
\small
\centering
  \begin{tabular}{p{0.225\textwidth}|p{0.135\textwidth}|p{0.115\textwidth}|p{0.04\textwidth}|p{0.08\textwidth}|l|p{0.05\textwidth}|l}
    \toprule
    \multirow{2}{*}{Dataset} & \multirow{2}{*}{Creation} & \multirow{2}{*}{KC}  & \multicolumn{2}{c|}{Question Types} & \multirow{2}{*}{PAT} & \multirow{2}{*}{\makecell{Auto.\\Ans-\\update}} & \multirow{2}{*}{\#ques.}\\
    \cmidrule(r){4-5}
    &  & & m-hop & Bef-event reasoning &  & &\\
    \midrule
    \multicolumn{2}{c}{\textbf{Temporal QA \& Reasoning Datasets}}& & & & &\\
    \midrule
    TempQuestions \citeyearpar{jia2018tempquestions} & Man.-Filt. & Freebase & \cmark & \cmark &  \xmark  & \xmark & 1271 \\
    CRON-QUESTIONS \citeyearpar{saxena2021question} & Templ. & Wikidata  & \cmark & \xmark &  \xmark  & \xmark &  410k\\
    TimeQA \citeyearpar{chen2021dataset} & Templ.- Wikidata & Wikipedia  & \xmark & \xmark &  \xmark & \xmark &  20k\\
    SituatedQA-temporal \citeyearpar{zhang2021situatedqa}  & Man.-Filt. & Wikipedia  & \xmark & \xmark &  \xmark & \xmark &  12k\\
    TEMPLAMA \citeyearpar{dhingra2022time}  & Templ./ Cloze & Custom-News & \xmark & \xmark &  \xmark  & \xmark &  50k \\
    StreamingQA \citeyearpar{liska2022streamingqa}  & Man.+Gen & WMT news &\cmark & \xmark &  \xmark  & \xmark & 410k\\
    TEMPREASON \citeyearpar{tan2023towards} & Templ./ Cloze & Wikidata  & \xmark & \cmark &  \xmark  & \xmark & 429k \\
    \midrule
    \multicolumn{2}{c}{\textbf{Present-Anchored Temporal QA Datasets}}& & & & &\\
    \midrule
    REALTIME QA \citeyearpar{kasai2022realtime} & News websites & News Articles  & \cmark & \xmark &  some & \xmark & $\sim$ 5k\\
    FreshQA \citeyearpar{vu2023freshllms} & Man. & Google search & \cmark & \xmark & 377 & \xmark &  600\\
    \textbf{PAT-Questions} (ours) & Templ.-Wikidata & Wikipedia &  \cmark & \cmark &  \cmark & \cmark &  6172\\
  \bottomrule
\end{tabular}
\caption{Comparison of temporal question-answering datasets. Abbreviations: Man.=created manually, Man.-Filt.=filtered from other datasets, Man.+Gen.=created by crowdsourcing and generated by LLMs, Templ.=created using templates, KC=Knowledge Corpus, PAT=Present Time-Anchored.}
\label{tab:datasetcomparison}
\end{table*}

Our main contributions are:
\vspace{-4pt}
\begin{itemize}
    %\item We introduce the problem of Present-anchored Temporal Question-answering (PAT-QA).
    \item We publish a novel PATQA benchmark, PAT-Questions\footnote{Our dataset and code for self-updates: \url{https://github.com/jannatmeem95/PAT-Questions.git}}, with annotated single-hop and multi-hop questions for two different timestamps (December 2021, December 2023). We provide an automatic answer updating system for the research community to always get up-to-date answers to PAT-Questions.\vspace{-3pt}
    % \item We modify a a state-of-the-art temporal QA system, \citep{tan2023towards}'s, which requires context as input, to work for open-retrieval PAT-Questions.
    \item We evaluate our benchmark on a wide range of LLMs in direct prompting and RAG settings, and identify limitations of the LLMs in tackling PAT-Questions. 
    \item We modify a state-of-the-art temporal reasoning system, \citet{tan2023towards}, to answer our PAT-Questions, and experimentally show how it performs on our dataset.
\end{itemize}
\vspace{-8pt}
\section{Related Work}
\paragraph{Temporal Question-Answering Datasets}
Research on understanding time in texts has led to the development of datasets aimed at enhancing temporal understanding in both knowledge-base question answering (KBQA) and natural language question-answering systems. Prior works on temporal KBQA have led to the creation of datasets like TempQuestions \citep{jia2018tempquestions}, Tequila \citep{jia2018tequila}, TimeQuestions \citep{jia2021complex}, and CRONQuestions \citep{saxena2021question}, which
focuses on integrating temporal data into knowledge bases for ranking entities related to a query \citep{talukdar2012coupled, chang2012sutime}. Recent efforts have shifted towards enhancing large language models (LLMs) for time-sensitive reasoning based on natural text only. Datasets like TimeQA \citep{chen2021dataset}, TEMPLAMA \citep{dhingra2022time}, and TEMPREASON \citep{tan2023can} have been introduced to test the ability of LLMs to reason and answer questions that involve understanding explicit temporal context (i.e. `What team did Cristiano Ronaldo play for in 2021?') or complex temporal relations such as `before' and `after' (i.e. `What team did Cristiano Ronaldo play for before Manchester United?') or to identify time-dependent facts from unstructured text.

\paragraph{Time-sensitive reasoning over Evolving data}
Existing benchmarks in temporal QA systems focus on static knowledge, annotating questions with single or explicit timestamps, which overlooks the dynamic nature of real-world information where answers can change over time. Notably, SituatedQA-temporal \citep{zhang2021situatedqa} and StreamingQA \citep{liska2022streamingqa} have attempted to incorporate temporal context by dating questions and sourcing from recent news, yet they still operate on static snapshots of knowledge. The dynamic REALTIME QA benchmark \citep{kasai2022realtime} tests models on current events, however, they exclusively focus on news data and lack emphasis on evolving facts and multi-hop reasoning. FreshQA \citep{vu2023freshllms} is a contemporary dataset that attempts to update LLMs with current information through time-sensitive questions. Both REALTIME QA and FreshQA rely on the authors to update the answers to reflect new information or changes over time, which limits the datasets' effectiveness in supporting the real-time adaptation of LLMs. In contrast, our dataset, PAT-Questions, can be automatically updated over time ensuring its adaptability and accuracy in real-time, surpassing existing present-anchored datasets. 
% PAT-Questions also introduce a comprehensive challenge by including 6172 questions that test LLMs on temporal understanding, multi-hop, and complex temporal relation reasoning, filling a significant gap left by previous datasets in assessing all three critical reasoning dimensions.
Table \ref{tab:datasetcomparison} shows the comparison among all relevant datasets.

\section{PAT-Questions Dataset Construction}
\begin{figure*}[t]
    \centering
    \includegraphics[width=0.85\textwidth]{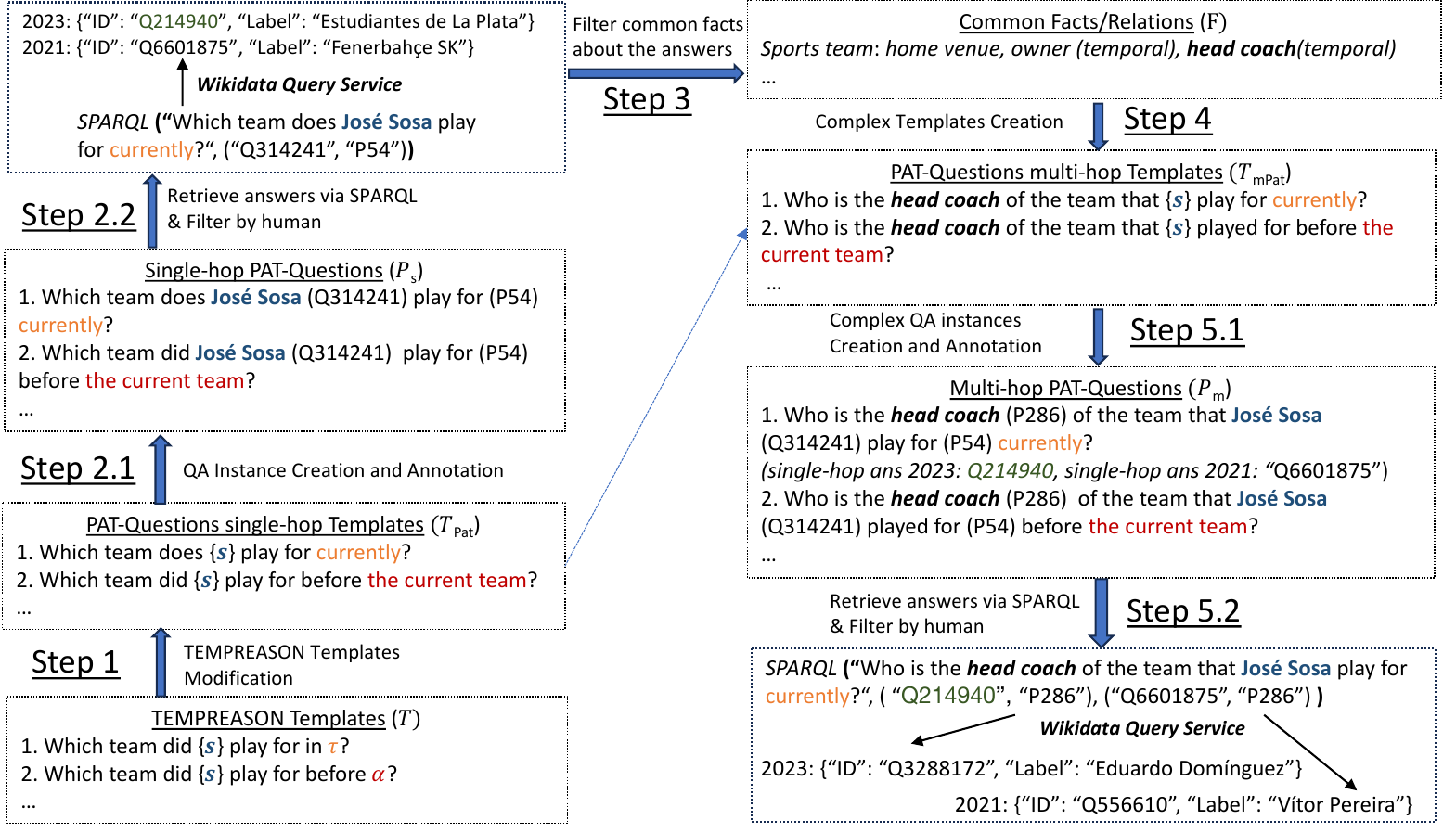}
    \caption{Illustration of PAT-Questions dataset construction following Algorithm \ref{alg:data_creation}. Firstly, we modify the time-sensitive templates from the TEMPREASON dataset \citep{tan2023towards} to build PAT-Questions templates, and following the steps shown in the figure, we create a set of one-hop and multi-hop PAT-Questions with annotated answers for two different timestamps, Dec 2021 and Dec 2023. Here, $\tau$  and $\alpha$ refer to a year and an entity respectively.}
    \label{fig:data_collections}
\end{figure*}
We extend the TEMPREASON dataset \citep{tan2023towards} to construct PAT-Questions. Each question in TEMPREASON follows a time-sensitive template and is annotated with Wikidata IDs for the primary subject and relation, which facilitates us to generate structured SPARQL queries to automatically update the responses over time. The pre-existing annotations eliminate the need for entity linking as a pre-processing step.
%We selected TEMPREASON over other temporal datasets because it is the most recently published temporal questions dataset built from a reliable and structured data source (Wikidata) and contains the most recent facts that are likely to vary over time. 
% Specifically, each question follows a time-sensitive template and is annotated with Wikidata IDs for the primary subject and relation, simplifying the retrieval of one-hop answers through SPARQL queries, and eliminating the need for entity linking as a pre-processing step. 
We leverage the, the December 31, 2023 Wikidata dump as the knowledge source. We annotate PAT-Questions for two different timestamps of Wikidata to compare the performance of the LLMs. The overall procedure of our data construction is illustrated in Figure \ref{fig:data_collections} and formally defined in Algorithm \ref{alg:data_creation}.

\textbf{Step 1: TEMPREASON Templates Modification}
TEMPREASON templates by \citet{dhingra2022time}, consist of time-sensitive facts ($s,r,o,\tau_s,\tau_e)$, with $s$ representing the subject, $r$ the relation, $o$ the object, and $\tau_s$ and $\tau_e$ denoting the start and end times of the fact. We adapt two types of TEMPREASON templates ($T$): i) $(s,r,?,\tau)$ (where $\tau$ lies between $\tau_s$ and $\tau_e$) becomes our single-hop PAT template $(s,r,?,\tau_{cur})$ where $\tau_{cur}$ is the current time, and ii) $(s,r,?,\tau\prec\tau_\alpha)$ (where $\alpha$ is an object related to ($s,r$) pair facts with distinct $\tau_s$ and $\tau_e$, and $\tau\prec\tau_\alpha$ is the time range immediately preceding $\tau_s$) transforms into $(s,r,?, \tau \prec \tau_{\alpha_{cur}})$ where $\tau \prec \tau_{\alpha_{cur}}$ represents the time range immediately preceding the start time of the current object of the ($s,r$) pair fact. These rules are outlined formally in Table \ref{tab:template_construction_rules}. Our templates ($T_{Pat}$) are challenging as they don't explicitly specify the current time $\tau_{cur}$ or object $\alpha$, unlike the original TEMPREASON templates. Steps 1-7 of Algorithm \ref{alg:data_creation} depict Step 1, illustrated with examples in Figure \ref{fig:data_collections}. Our PAT-Questions single-hop templates are available in Table \ref{tab:simpletemplates} in Appendix \ref{app:template_details}.
% TEMPREASON templates are built on ten time-sensitive facts following the work by \citet{dhingra2022time}. A time-sensitive fact is represented as ($s,r,o,\tau_s,\tau_e)$, where $s$ is the subject, $r$ is the relation, $o$ is the object, $\tau_s$ and $\tau_e$ are the start time and end time of this fact.
% We modify two types of TEMPREASON templates ($T$): i) $(s,r,?,\tau)$ (where $\tau$ is a time between $\tau_s$ and $\tau_e$) to our single-hop PAT template $(s,r,?,\tau_{cur})$ where $\tau_{cur}$ is the present time, and ii) $(s,r,?,\tau\prec\tau_\alpha)$ (where $\alpha$ is an object of one of the ($s,r$) pair facts for a distinct $\tau_s$ and $\tau_e$, and $\tau\prec\tau_\alpha$ is the time range immediately preceding $\tau_s$) to $(s,r,?, \tau \prec \tau_{\alpha_{cur}})$ where $\tau \prec \tau_{\alpha_{cur}}$ represents the time range immediately preceding the starting time of the current or present object of the ($s,r$) pair fact. We show the rules formally in Table \ref{tab:template_construction_rules}. Our templates ($T_{Pat}$) are challenging because explicit time, $\tau$, and object $\alpha$ for the current time $\tau_{cur}$, both are not explicitly specified, unlike the TEMPREASON templates. Lines 1-7 of Algorithm \ref{alg:data_creation} represent Step 1 and the illustration is shown with a few examples in Figure \ref{fig:data_collections}. The original templates from the TEMPREASON dataset \citep{tan2023towards} and our modified PAT-Questions single-hop templates can be found in Table \ref{tab:simpletemplates} in Appendix \ref{sec:appendix}.
\begin{algorithm}[t]
\algsetup{linenosize=\tiny}
\small
\caption{Construct PAT-Questions Dataset}
\begin{algorithmic}[1]
% \small
\label{alg:data_creation}
\REQUIRE TEMPREASON dataset, $D$, TEMPREASON templates $T$
\ENSURE PAT-Questions\\
\STATE $T_{Pat}$ $\leftarrow []$ $\backslash\backslash$ Single-hop templates
%($s,r,\tau/\alpha)
\FOR{\textit{each template}, $t \in T$}
    \STATE $\backslash\backslash t = (s,r,\tau)$, or $t = (s,r,\alpha)$
    \IF{$\tau \in t$}
        \STATE $T_{Pat}$ $ \leftarrow$ \textit{Replace}($\tau$, `currently') $\cup T_{Pat}$
    \ELSIF{$\alpha \in t$}
        \STATE $T_{Pat} \leftarrow$ \textit{Replace}($\alpha$, `current'+\textit{equiv($r$)}) $\cup T_{Pat}$
    \ENDIF
    \COMMENT{$~using~rules~from~Table~$\ref{tab:simpletemplates}}
\ENDFOR
\STATE $S$ =  [subjects($D$)]\\ $\backslash\backslash$ Wikidata subjects for all TEMPREASON questions
\STATE $P_s =$ \textit{CreateQAInstances}(\textit{SPARQL}($S, T_{Pat}))$)\\
\STATE  $F = $ {\textit{Filter(multiFacts}}($P_s$))\\
\STATE $T_{mPat} \leftarrow []$ $\backslash\backslash$ Multi-hop templates
\FOR{\textit{each relation}  $r_i \in F$} 
    \STATE $T_{mPat}\leftarrow$ \textit{Insert}($r_i, T_{Pat}$)$~\cup T_{mPat}$
    \COMMENT{$~using~rules~from~Table~$\ref{tab:complextemplates}}
\ENDFOR
\STATE $P_m =$\textit{CreateQAInstances}(\textit{SPARQL}($S, T_{mPat}))$
\STATE PAT-Questions $=P_s \cup P_m$
\RETURN PAT-Questions
\end{algorithmic}
\end{algorithm}

\textbf{Step 2: Simple QA instances Creation and Annotation}
\begin{table*}[t]
\small
    \centering
    \begin{tabular}{p{0.1\textwidth}|p{0.37\textwidth}|p{0.2\textwidth}|p{0.23\textwidth}}
        \toprule
        \textbf{KB relation, $r$} & \textbf{Rule} & \textbf{TEMPREASON Template} & \textbf{PAT-Questions single-hop Template}\\
        \midrule
        member of sports & $(s,r,?,\tau) \rightarrow (s,r,?,\tau_{cur})$ & Which team did \{$s$\} play for in $\bm{\tau}$?  & Which team does \{$s$\} play for \textbf{currently}?\\
        \cmidrule(r){2-4}
        team (P54) & $(s,r,?,\tau\prec\tau_\alpha) \rightarrow (s,r,?, \tau \prec \tau_{\alpha_{cur}})$ & Which team did \{$s$\} play for before $\bm{\alpha}$?  & Which team did \{$s$\} play for before \textbf{the current team}? \\
        \bottomrule
    \end{tabular}
    \caption{Conversion of the TEMPREASON templates (Step 1) for the `member of sports team' relation to single-hop PAT-Question templates. TEMPREASON has two templates per relation $r$ and we convert each of the templates following the two rules shown above. For example, $s$ is \textit{Cristiano Ronaldo}, $\tau$ is $2021$ and $\alpha$ is \textit{Real Madrid F.C.}, the single-hop PAT-Questions become `Which team does \textit{Cristiano Ronaldo} play for \textit{currently}?' and  `Which team did \textit{Cristiano Ronaldo} play for before \textit{the current team}?'}
    \label{tab:template_construction_rules}
\end{table*}
\begin{table*}[t]
\small
    \centering
    \begin{tabular}{p{0.08\textwidth}|p{0.12\textwidth}|p{0.21\textwidth}|p{0.185\textwidth}|p{0.27\textwidth}}
        \toprule
        \textbf{KB relation, $r$} & \textbf{Common relations, $r_i$} &\textbf{Rule} & \textbf{PAT-Questions single-hop Template} & \textbf{PAT-Questions multi-hop Template}\\
        \midrule
        member of sports & home venue (P115), & $(s,r,?,\tau_{cur}) \rightarrow ((s,r,?,\tau_{cur}), r_i, ?, \tau_{cur})$ & Which team does \{$s$\} play for currently & What is the \textbf{home venue} of the team that \{$s$\} plays for currently?\\
        team (P54) & head coach (286) & & $(\tau_{cur})$? & Who is the \textbf{head coach} of the team that \{$s$\} plays for currently?\\
        \cmidrule(r){3-5}
         &  & $(s,r,?, \tau \prec \tau_{\alpha_{cur}}) \rightarrow ((s,r,?, \tau\prec\tau_{\alpha_{cur}}), r_i,$ $?, \tau_{cur})$ & Which team did \{$s$\} play for before the current team $(\tau \prec\tau_{\alpha_{cur}})$? & What is the \textbf{home venue} of the team that \{$s$\} played for before the current team? \\
         & & &  &Who is the \textbf{head coach} of the team that \{$s$\} played for before the current team?\\
        \bottomrule
    \end{tabular}
    \caption{Conversion of the PAT-Questions single-hop templates to multi-hop templates (Step 4) for the `member of sports team' (P54) relation to PAT-Questions multi-hop templates.}
    \label{tab:multi_hop_template_construction_rules}
\end{table*}
We filter the subject entities from original TEMPREASON questions for which the PAT-Questions are valid. Based on \citet{tan2023towards}'s approach, we insert the subjects into the single-hop PAT-Questions templates ($T_{Pat}$) and annotate the questions with the Wikidata IDs of the subjects and relations.  Since questions are in natural language, we establish a set of SPARQL query templates to convert each natural language question into its corresponding SPARQL query (see Appendix \ref{app:data_stat}). We insert each ($s,r$) pair into the appropriate SPARQL template, and retrieve the Wikidata ID and NL label of the gold answer using the Wikidata Query Service API(Algorithm \ref{alg:data_creation}, lines 8-9). Note that questions are annotated for two different timestamps. We temporally organize the facts linked with ($s,r$) pairs, fetching the latest objects ($o$) (current and previous) for the 2023 version, and filtering by end date $\tau_e \leq Dec, 2021$ for the 2021 version.

% For each query, we use Wikidata Query Service API to fetch the Wikidata ID and natural language label of the answer to the query. As the questions are in natural language (NL), we define a set of SPARQL query templates to convert each NL question to its corresponding SPARQL query (Appendix \ref{fig:single-sparql}). As mentioned earlier, we annotate the questions for two different timestamps. We temporally sort the facts associated with the ($s,r$) pairs, and fetch the latest objects ($o$) (current and before-current) for annotating 2023 version, and filter by end date $\tau_e \leq 2021$ for annotating 2021 version. We insert each ($s,r$) pair to the appropriate SPARQL template and retrieve Wikidata ID and NL label of the gold answer (Alg. \ref{alg:data_creation} lines 8-9). We filter out the questions whose answers could not be found or retrieved by SPARQL query.
\textbf{Step 3: Filter common facts}
In this step, we randomly select a subset of single-hop question-answer pairs, capturing all templates. We extract all facts ($F$), including both temporal and non-temporal ones, linked to the answer Wikidata entities (Algorithm \ref{alg:data_creation}, line 10). We then filter common facts, i.e., Wikidata triples ($s,r,o$) shared among these entities, which can be temporal or static. Notably, we prioritize single-hop answer facts over subjects to construct multi-hop templates. This decision is made because single-hop answers span various types, such as sports teams, employers, heads of government/company/organization, etc., resulting in a broader range of facts for our multi-hop questions compared to those associated with the subjects.

% In this step, we first choose a random subset of the single-hop question-answer pairs covering all single-hop templates and extract all facts ($F$), both temporal and non-temporal, associated with the answer Wikidata entities (Alg. \ref{alg:data_creation} line 10). We filter the common facts, meaning Wikidata triples ($s,r,o$) that exist for all these entities. These common facts may either be temporal or static. Note that we chose the facts of the single-hop answers instead of the subjects to build the multi-hop templates because the single-hop answers are of different types such as sports teams, employers, heads of the government/company/organization, etc., and the facts associated to these types vary a lot in comparison to the facts associated with the subjects ensuring a varied range of facts in our multi-hop questions.
\textbf{Step 4: Complex Templates Creation}
We generate multi-hop PAT-Question templates ($T_{mPat}$) by integrating facts from $F$ into the single-hop templates, $T_{Pat}$, and converting them into natural language following the guidelines outlined in Table \ref{tab:multi_hop_template_construction_rules}, where $r_i$ represents one of the relations in $F$ (Algorithm \ref{alg:data_creation}, lines 11-13). Note that all answers to the multi-hop templates are grounded on the time the question is posed, denoted as $\tau_{cur}$. The multi-hop PAT templates are listed in Table \ref{tab:complextemplates} in Appendix \ref{app:template_details}.
% We create multi-hop PAT-Question templates ($T_{mPat}$) by inserting facts from $F$ into the single-hop templates, $T_{Pat}$ and converting them to natural language, following the rules mentioned in Table \ref{tab:multi_hop_template_construction_rules}, where $r_i$ is one of the relations in $F$ (Alg. \ref{alg:data_creation} lines 11-13). Note that all the answers to the multi-hop templates are grounded in the time the question is asked $\tau_{cur}$. The multi-hop PAT templates can be found in Table \ref{tab:complextemplates} in Appendix \ref{sec:appendix}. 
% \yue{can you write an alg for this data construction?} \meem{Done}
\begin{figure*}[t]
    \centering
    \includegraphics[width=1\textwidth]{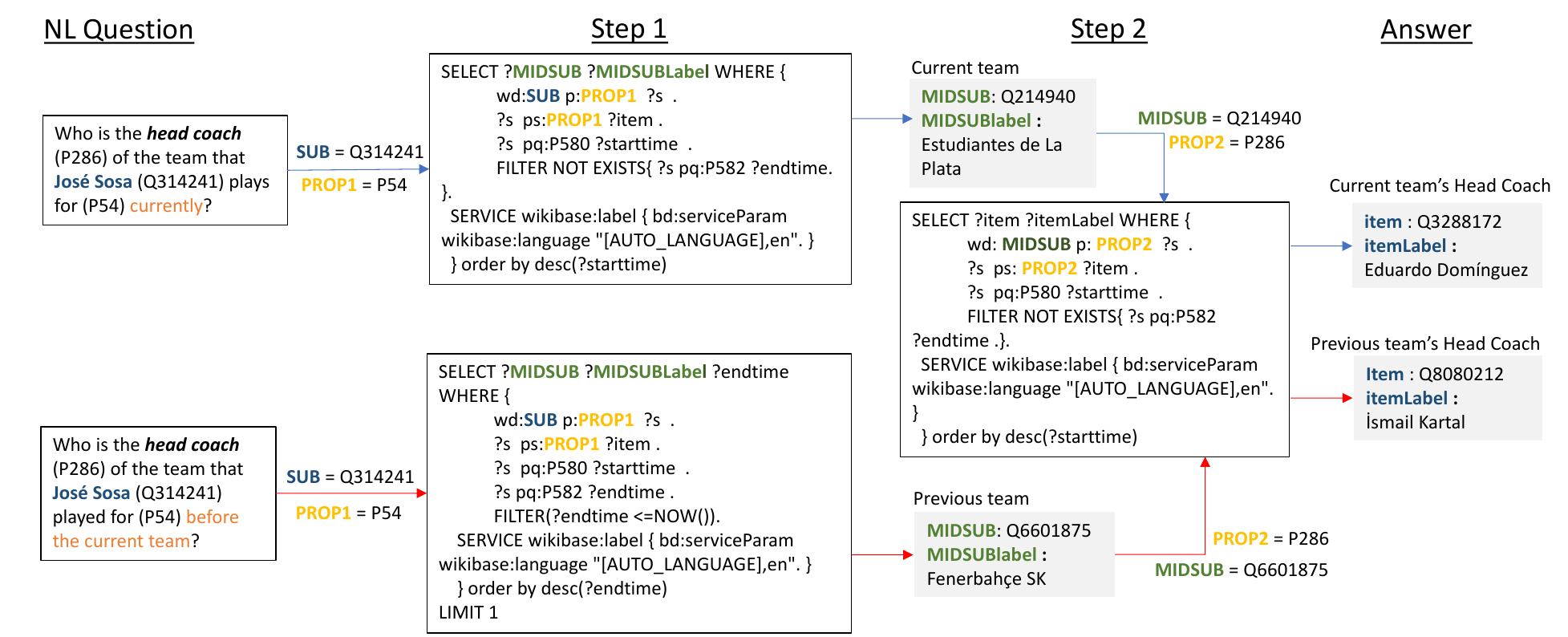}
    \caption{Illustration of automatic answer-updates to two multi-hop PAT-Questions via SPARQL templates}
    \label{fig:sparl_ans_update}
\end{figure*}

\textbf{Step 5: Complex QA instances Creation and Annotation}
For each filtered question in step 2, we insert the subject into its multi-hop template (Algorithm \ref{alg:data_creation}, line 14), annotating it with the subject, relation, intermediate entity (gold answer to the single-hop question), and intermediate relation ($r_i$). Answers are then retrieved following step 3. In this process, we select intermediate entities and relations for insertion into SPARQL templates for most of the questions (see Appendix \ref{app:sparql}), rather than the original subject and relation. Some questions are filtered out due to missing facts, like spouse (P26), founder (P112), etc. Detailed statistics for \textbf{PAT-Questions} are provided in Table \ref{tab:PAT-Questions_stat}.

% For each question filtered and annotated in step 2, we insert the subject to its corresponding multi-hop template (Algorithm \ref{alg:data_creation} line 14). We annotate each question with the subject, relation, intermediate entity (gold answer to the single-hop question), and intermediate relation ($r_i$) in this step. We retrieve the answers following step 3. The only difference is that for this step, we pick the intermediate entity and intermediate relation to insert into the SPARQL templates (Appendix \ref{fig:multi-sparql}) instead of the original subject and relation. As we pick the common facts based on a subset of the questions, some facts (such as spouse (P26), founder (P112), etc.) may not be present for some entities. We filter out such questions. The statistics of \textbf{PAT-Questions} can be found in Table \ref{tab:PAT-Questions_stat}. 
We randomly select 1000 PAT-Questions and manually verify the accuracy of the annotated answers. We also filter out any questions where the answer cannot be retrieved via the SPARQL query.
% Note that we pick a random sample of 1000 questions from the dataset and manually check and confirm that the correctness of the annotated answers. We also filter out any question for each the answer cannot be retrieved by the SPARQL query.

\begin{table}[h]
\small
    \centering
    \begin{tabular}{c|c|c}
    \toprule
         \multirow{2}{*}{Type}& \multicolumn{2}{c}{Category}  \\
         \cmidrule(r){2-3}
         & single-hop & multi-hop\\
         \midrule
         current & 1442 & 1617 \\
         before-current & 1440 & 1673\\
         \midrule
         {\textbf{Total \# questions}} & 2882 & 3290\\
         \cmidrule(r){2-3}
         & \multicolumn{2}{c}{6172}\\
         % \midrule
         % {\textbf{Total \# questions }} & 617  & 508\\
         % \cmidrule(r){2-3}
         % \textbf{with different answer} & \multicolumn{2}{c}{}\\
         % \textbf{for 2021 and 2023} & \multicolumn{2}{c}{1125}\\
         \bottomrule
    \end{tabular}
    \caption{Dataset Statistics of PAT-Questions}
    \label{tab:PAT-Questions_stat}
\end{table}
\vspace{-12pt}
\paragraph{Automatically Updating the Answers of PAT-Questions.}
The questions in our dataset are time-sensitive, with answers expected to change periodically. While the most recent object of Wikidata facts may change, the subject and relation remain constant (Example in Figure~\ref{fig:intro}(a)). Thus, the SPARQL template associated with each question consistently retrieves the latest answer without requiring manual intervention. This functionality empowers users to update the answers to PAT-Questions any time they want. An illustration of the answer update process is provided in Figure \ref{fig:sparl_ans_update}.
% The questions within our dataset are time-sensitive, with answers expected to change periodically. 
Most questions include facts prone to change every six months or longer. To ensure that the research community has the latest answers, we commit to quarterly updates each year, executed through a cronjob running SPARQL queries automatically.
\section{Experiments}
\begin{table*}[t]
\small
    \centering
    \begin{tabular}{p{0.24\textwidth}|p{0.05\textwidth}|p{0.05\textwidth}|p{0.05\textwidth}|p{0.05\textwidth}|p{0.05\textwidth}|p{0.05\textwidth}|p{0.05\textwidth}|p{0.05\textwidth}}
    \toprule
    \multirow{2}{*}{} & \multicolumn{4}{c|}{Single-hop} & \multicolumn{4}{c}{Multi-hop}\\
    \cmidrule(r){2-9}
          &  \multicolumn{2}{c|}{2023} & \multicolumn{2}{c|}{2021} &  \multicolumn{2}{c|}{2023} & \multicolumn{2}{c}{2021}\\
          \cmidrule(r){2-9}
          & EM & F1 & EM & F1  & EM & F1  & EM & F1\\
          \midrule
          Falcon-7B & 4.4 & 5.7 & 7.8 & 5.8 & 2.5 & 5.6 & 4.4 & 6.5\\
          Falcon-7B-w-RAG & 8.1 & 4.9 & - & - & 4.7 & 2.9 & - & -\\
        \midrule
         Flan-T5-XL & 2.0 & 5.5 &  2.1 & 6.0 & 1.5 & 5.4 & 2.8 & 9.7\\
         Flan-T5-XL-w-RAG & 14.9 & 15.8 & - & - & 5.1 & \textbf{9.5} & - & -\\
         \midrule
         Llama-2-7B & 8.4 & 9.0 & 10.0 & 11.2 & 5.3 & \textbf{8.6} & 7.0 & 9.6\\
         Llama-2-7B-w-RAG & 13.9 & 8.7 & - & - & 6.6 & 6.0 & - & -\\
         \midrule
         Mistral-7B & 7.4  & 6.4 & 10.5 & 7.5 & 5.7 & 4.7 & 6.1 & 4.8 \\
         Mistral-7B-w-RAG & 12.7 & 5.5 & - & - & 5.9 & 2.7 & - & -\\
         \midrule
         GPT-3.5 & \textbf{11.7} & 11.3 & 13.6 & 13.3 & \textbf{9.3} & 7.7 & 9.7 & 8.1\\
         GPT-3.5-w-RAG & \textbf{15.5} & \textbf{16.5} & - & - & \textbf{7.6} & 6.6 & - & -\\
         \midrule
         TEMPREASON-T5-subWiki & 12.0 & \textbf{21.4} & -  & - &  2.3  & 7.9 &  - & - \\
         \midrule
         TEMPREASON-T5-w-RAG & 8.3 & 16.1 & - & - & 1.5 & 5.5 & - & -\\
         \bottomrule
    \end{tabular}
    \caption{The experimental results by EM Accuracy (\%) and token-level F1 (\%), for two categories of questions of PAT-Questions for two different snapshots of present data (Dec 2023 and Dec 2021)}
    \label{tab:baseline_experiments}
\end{table*}
\vspace{-4pt}
We conduct experiments on 5 LLMs that have been significantly successful in QA tasks but do not have access to up-to-date world knowledge, including Falcon-7B-Instruct \citep{falcon}, Flan-T5-XL \citep{chung2022scaling}, Llama-2-7B \citep{touvron2023llama}, Mistral-7B \citep{jiang2023mistral}, and GPT-3.5 \citep{brown2020language} in a direct prompting setting and a RAG setting. We also modify the existing setting of TEMPREASON-T5-SFT by \citet{tan2023towards} to evaluate PAT-Questions in a RAG setting.
We compare the results of direct prompting setting at two different timestamps: December 2021 and December 2023. Given that the cutoff date of the LLMs' knowledge is $\ge 2021$, they should ideally know the answers for December 2021. 
%In this section, we evaluate different LLM performances on our PAT-Questions dataset in direct prompting and RAG settings, comparing results for two timestamps: December 2021 and December 2023. This comparison is crucial given the cutoff of the LLMs' knowledge is $>= 2021$, they should ideally know answers to December 2021 annotations. \meem{add tempreason}
% In this section, we evaluate the performances of various LLMs on our PAT-Questions dataset in direct prompting and RAG settings, analyze and discuss the results. We compare the results of the LLMs for two different timestamps, December 2021 and December 2023 because most of the parametric knowledge of the open-source LLMs is cut off around or after the end of 2021, meaning the LLMs should ideally know the answers to the questions concerning the December 2021 gold annotations.
\vspace{-6pt}
\subsection{Experimental Setup}
\vspace{-6pt}
\textbf{Directly Prompting the Pre-trained Models}
In this experimental setting, we feed each question to the LLMs and instruct the LLMs to answer the question in a few words to avoid verbosity for EM comparisons (see Section \ref{sec:evaluation_metric}). We use the HuggingFace library for the open-source models and GPT-3.5-Chat API with a temperature of 0. For the 2021 evaluation of the open-source models, we prepend the question with ``Assume it is now December 2021," to ensure the fairness of the comparisons with the 2021 gold annotations and with GPT-3.5 for which the cutoff date is January 2022. 
% As expected, LLMs perform poorly for the questions requiring up-to-date knowledge, i.e. Dec 2023 annotations, however, they do not perform well on the Dec 2021 annotation as well. 
\\
\textbf{Retrieval-Augmented Generation (RAG)} \label{sec:obqa}
In this setting, we augment the LLMs' answer-generation capabilities with retrieval. We retrieve up to five Wikipedia documents for each question using Google Custom Search (GCS) Engine \footnote{\url{https://programmablesearchengine.google.com/}}, divide each document into chunks of 300 tokens, rank the relevance of these chunks using BM25 and finally assign the top 5 chunks as the retrieved evidence for a question from PAT-Questions dataset. Chunking is necessary in our case because the LLMs that we use have token limitations. We retrieve all the documents for all the questions on the same date (January 16, 2024) to maintain the fairness of our evaluation on the entire dataset. 
We prompt the LLMs using the question and the retrieved chunks and instruct the LLMs to answer in a few words using the information available in the chunks. We exclusively evaluate this method against December 2023 gold annotations since the retrieved documents contain current information. It would be illogical to retrieve data from a current knowledge source and compare it with outdated gold answers.
% We only compare the results of this setting with December 2023 gold annotations as the retrieved documents contain up-to-date information and it does not make sense to retrieve from an up-to-date knowledge source and expect an outdated answer.
\begin{figure*}[t]
    \begin{subfigure}[b]{0.5\textwidth}
      \centering
      \includegraphics[width=0.85\linewidth]{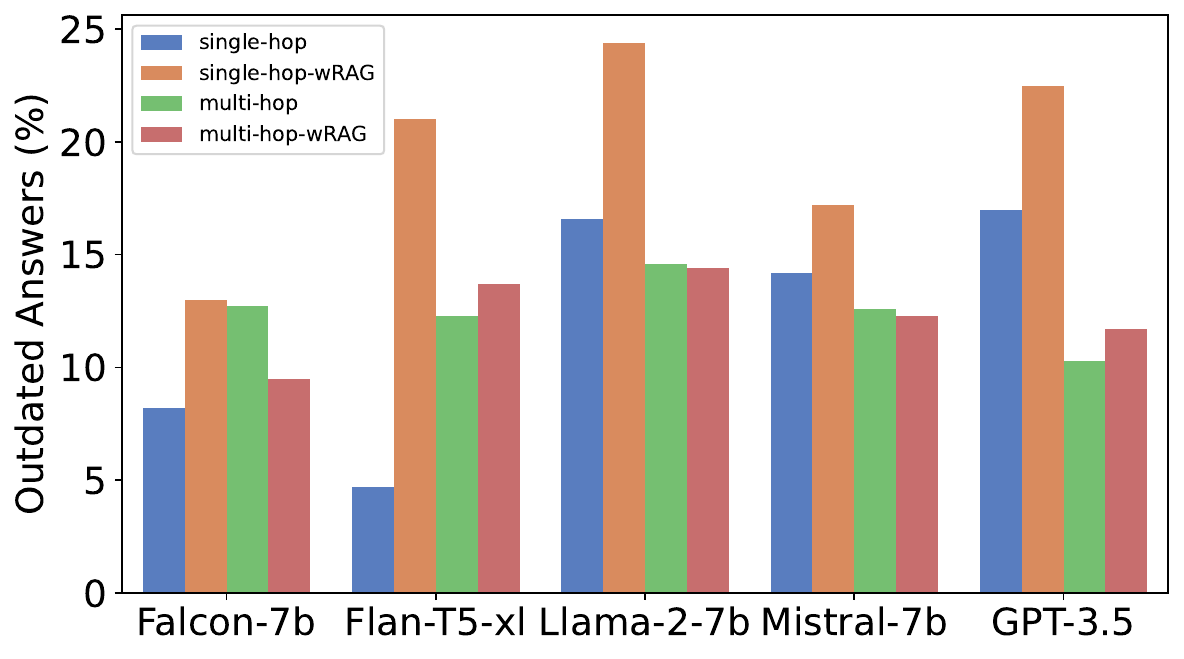}
      \caption{outdated responses (\%) by LLMs }
      \label{fig:outdated answer}
    \end{subfigure}%
    ~
    \begin{subfigure}[b]{0.5\textwidth}
      \centering
      \includegraphics[width=0.85\linewidth]{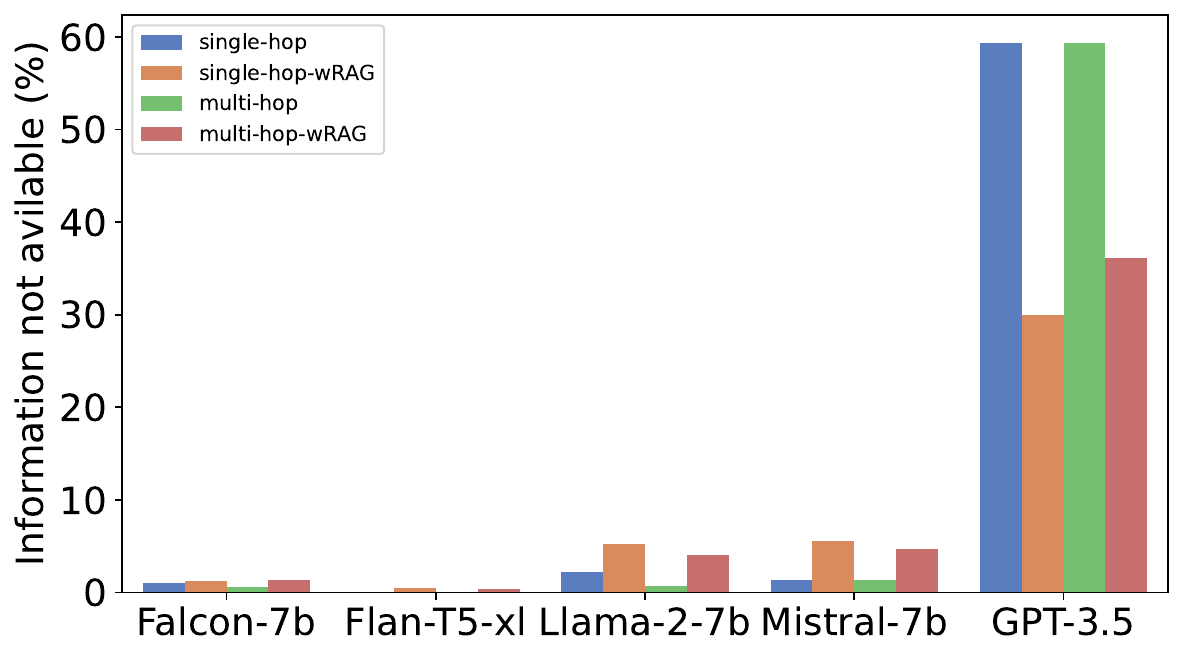}
      \caption{`information not available' responses (\%) by LLMs}
      \label{fig:notAvailable}
    \end{subfigure}%
\caption{Error distribution of the incorrect LLM responses}
    \label{fig:error_analysis}
\end{figure*}
\\
\textbf{TEMPREASON-T5 Experiments}
We evaluate our PAT-Questions with \citet{tan2023can}' T5-SFT model fine-tuned on improving the reasoning capability of the large language model by temporal span extraction. 
% In their work, they experimented with three settings, Closed-Book QA, Open-Book QA, and ReasonQA. In the ReasonQA setting, they provide gold temporal context extracted from Wikidata to show how well the model can reason over the temporal expressions. However, our questions contain multi-hop and complex temporal relations, and the ReasonQA setting is unsuitable for our evaluation. In
Their Open-book QA (OBQA) setting assumes that the subject entity of the question is already known and they extract the Wikipedia page associated with the subject entity to provide as context to the model. However, this setting is not practical in traditional Open-Retrieval QA settings. As such, we modify their OBQA setting to suit the PATQA problem. 
We provide the Top 5 BM25 Wikipedia chunks retrieved by GCS for the question as context and, and evaluate their fine-tuned model's performance (TEMPREASON-T5-w-RAG. in Table \ref{tab:baseline_experiments}). We also show a comparison with their version of the OBQA setting, meaning we extract the content of the subject entity's current Wikipedia page and provide that as context to the model (TEMPREASON-T5-subWiki in Table \ref{tab:baseline_experiments}).
\paragraph{Evaluation Metrics}\label{sec:evaluation_metric}
% We use Exact Match Accuracy and token-level F1 \citep{rajpurkar2016squad} to measure the correctness of the generated answers to the questions of PAT-Questions. 
We employ token-level F1 \citep{rajpurkar2016squad} and  \citet{chen2023benchmarking}'s exact matching (EM) Accuracy metric for the LLMs where if the generated text contains an exact match to the answer or vice-versa, it is considered a correct answer. To address the issue where LLMs might produce an accurate yet differently phrased response to PAT-Questions, such as "Man United" instead of "Manchester United F.C.," resulting in a zero exact match (EM) score, we annotate each answer with all possible aliases from Wikidata using SPARQL queries. 
% The answers to PAT-Questions are always Wikidata entities with a unique label or name. However, the LLMs may generate a paraphrased version or a part of the label. For example, if the gold label is `Manchester United F.C.', and the LLM generates `Man United', EM will be zero for this case although the LLM-generated answer is correct. To overcome this limitation, we annotate each answer with all the aliases or paraphrases of the answer entities extracted from the Wikidata KB (via SPARQL).
For \citet{tan2023towards}'s system, we use traditional Exact Match and F1 to be consistent with their evaluation.
\subsection{Results and Discussion} \label{sec:resuts}
Our findings, presented in Table \ref{tab:baseline_experiments}, indicate that pre-trained Large Language Models (LLMs) face challenges with PAT-Questions, both single and multi-hop, showing very low EM scores between 1.5\% to 15.5\% and F1 scores ranging from 2.9\% to 16.5\%.  Accuracy improves with document retrieval, especially for single-hop questions, due to the retrieval of up-to-date and relevant documents.
% We show our experimental results in Table \ref{tab:baseline_experiments}. All the pre-trained LLMs struggle to answer both single and multi-hop PAT-questions. Both the EM scores range from 1.5\% to 15.5\% and F1 scores are very low ranging from 2.9\% to ~16.5\% for both direct prompting and RAG settings. The accuracy increases for all the models with document retrieval (w-RAG in Table \ref{tab:baseline_experiments}),  especially for single-hop questions and the models provide better results for the 2023 annotation set as the documents retrieved are up-to-date and relevant. 
Open-source LLMs, significantly underperform in direct prompting settings compared to GPT-3.5 for multi-hop questions. These models considerably benefit from document retrieval due to their lower initial baseline. However, the success of the RAG approach largely depends on the retrieval engine's efficiency, which in our case struggles more with multi-hop than single-hop questions as evidenced by the performance degradation of GPT-3.5 for multi-hop questions. 
% However, GPT-3.5's superior performance degrades due to poor document retrieval for multi-hop questions. The performance of the RAG setting largely depends on the effectiveness of the retrieval engine and it is evident from the results that the retrieval performance of GCS is much worse for multi-hop questions compared to single-hop questions. 
% The success of the Retrieve-and-Generate (RAG) approach largely hinges on the retrieval engine's efficiency, which struggles more with multi-hop than single-hop questions. Despite post-2021 data cutoffs, LLM performance on both PAT and multi-hop reasoning remains low, underscoring the models' limitations.
Despite the LLMs' knowledge cut-off date being $\ge 2021$, the performance compared to 2021 annotations is still very low (though better than the up-to-date annotations). This highlights the LLMs' performance gap in both PAT and multi-hop reasoning. Note that the F1 scores for different models show considerable variation. Flan-T5-XL and GPT-3.5 generally adhere to instructions for concise responses, leading to brief and focused answers. Conversely, other models, including GPT-3.5 in certain instances, tend to produce longer responses, which, despite being accurate, result in lower F1 scores due to their verbosity.

We also compare the performances of TEMPREASON-T5 model with two different contexts: the subject's Wikipedia page and the documents retrieved by GCS. 
 % The performance of the model on PAT-Questions even before varying the context setting is relatively low (TEMPREASON-T5-subWiki), especially for multi-hop questions. 
Although the model is specialized in temporal reasoning on the subject's Wikipedia content, it shows low accuracy on both single and multi-hop PAT-Questions. However being fine-tuned on single-hop temporal facts from Wikidata, the model demonstrates comparable results with the open-source LLMs on single-hop questions. The performance degrades significantly for multi-hop questions and open-retrieval RAG settings due to the lack of multi-hop and PAT reasoning capabilities. 

We presented a random subset of 50 multi-hop PAT-Questions to New Bing and GPT-4 Web. New Bing accurately answered 9 questions but failed or provided incorrect responses for the remaining 41. GPT-4, on the other hand, correctly answered 6 questions, inaccurately responded to 6, and indicated that information was unavailable for the remaining 38 questions. This comparison highlights the challenges both the services face in handling multi-hop PAT-Questions (see Appendix \ref{app:prompts}). 
% The findings indicate significant difficulties faced by both LLMs (LLMs) and advanced reasoning APIs when addressing PAT-Questions, highlighting their limitations in complex question handling.
% These results prove that both the LLMs and the state-of-the-art temporal reasoning model greatly suffer in handling PAT-Questions. 

\textbf{Error Analysis} 
Figure \ref{fig:error_analysis} shows the error distribution of the LLM-generated answers. Figure \ref{fig:outdated answer} shows the percentage of outdated answers and Figure \ref{fig:notAvailable} shows the percentage of `information not available' or similar responses out of the incorrect responses of the LLMs based on EM. The responses of Llama-7b, Mistral-7b and GPT-3.5 (especially GPT-3.5) are more grounded to the information available in their parametric memory till the cut-off date for single-hop questions, whereas Flan-T5-XL and Falcon-7b are more likely to generate fake or misinformed responses when not prompted with RAG. Almost all the LLMs struggle with multi-hop reasoning. GPT-3.5 is more cautious in answering present-centric questions and is more likely to respond with `I do not have real-time information' than responding with an incorrect or outdated answer (see Appendix \ref{app:prompts} and \ref{filter_error} for more details).
\vspace{-2pt}
\subsection{Additional Experimental Results}
\textbf{   Direct Prompting without specifying date} We ran experiments on the open-source LLMs with direct prompting without specifying ``Assuming it is now December, 2021" in the prompt. The results comparing the generated responses with the December 2021 snapshot are shown in Table \ref{tab:withoutAssume}. The results are lower than that of when the date is specified i.e. the scores shown in Table \ref{tab:baseline_experiments}. We also compare the RAG results with 2021 gold annotations and find accuracies going up to some extent (in Table \ref{tab:2021_RAG}). This is due to some questions of our dataset still having the same answers as they were in December 2021, however, the LLMs could not correctly answer those. By integrating RAG, those answers have been correctly identified by the LLMs and TEMPREASON-T5.
\begin{table}[t]
\small
    \centering
    \begin{tabular}{p{0.1\textwidth}|p{0.05\textwidth}|p{0.05\textwidth}|p{0.05\textwidth}|p{0.05\textwidth}}
    \toprule
    \multirow{2}{*}{} & \multicolumn{2}{c|}{Single-hop} & \multicolumn{2}{c}{Multi-hop}\\
    \midrule
          & EM & F1 & EM & F1  \\
          \midrule
          Falcon-7B &  4.7 & 6.3  & 3.6 & 5.9\\
        \midrule
         Flan-T5-xl &  2.1 & 5.7  & 2.6 & 5.7\\
         \midrule
         Llama-2-7B &  9.8 & 10.2  & 6.0 & 9.0\\
         \midrule
         Mistral-7B &  8.6 & 7.0  & 6.1 & 4.7 \\
         \bottomrule
    \end{tabular}
    \caption{The experimental results by EM Accuracy (\%) and token-level F1 (\%), for two categories of questions of PAT-Questions for Dec 2021 snapshot without pretending ``Assuming it is December 2021" to the prompt.}
    \label{tab:withoutAssume}
\end{table}
\begin{table}[t]
\small
    \centering
    \begin{tabular}{c|c|c|c|c}
    \toprule
     & \multicolumn{2}{c|}{Single-hop} & \multicolumn{2}{c}{Multi-hop}\\
          \midrule
         & EM & F1  & EM & F1 \\
          \midrule
          Falcon-7B-w-RAG & 8.1 & 5.1 &  4.7 & 2.9\\
        \midrule
         Flan-T5-XL-w-RAG  & 12.9 & 14.5 & 5.7 & 9.7\\
         \midrule
         Llama-2-7B-w-RAG & 13.0 & 8.7 & 7.1 & 6.0\\
         \midrule
         Mistral-7B-w-RAG & 12.2 & 5.0 & 6.1 & 2.6\\
         \midrule
         GPT-3.5-w-RAG & 14.8 & 16.3  & 7.9 & 6.7\\
         \midrule
         TEMPREASON-T5-subWiki & 11.4  & 20.7 &  2.4 & 8.0 \\
         \midrule
         TEMPREASON-T5-w-RAG & 1.5 & 5.5 & 1.5 & 5.5\\
         \bottomrule
    \end{tabular}
    \caption{The experimental results for RAG setting by EM Accuracy (\%) and token-level F1 (\%), for two categories of questions of PAT-Questions for December 2021 timestamp}
    \label{tab:2021_RAG}
\end{table}
\begin{table}[t]
\small
    \centering
    \begin{tabular}{c|c|c}
    \toprule
    Method & LLM & EM (\%)\\
          \midrule
         Chain-of-Thought (CoT) & Llama-2-7B & 11.2 \\
         & GPT-3.5 & 16.0\\
         \midrule
         ReAct & GPT-3.5 & 8.8\\
         \midrule
         Verify-and-Edit & GPT-3.5 & 8.9\\
         \bottomrule
    \end{tabular}
    \caption{The EM Accuracy (\%), for the multi-hop PAT-Questions (Dec 2023) using advanced CoT prompting (\citet{wei2022chain}), and advanced RAG methods (\citet{yao2022react, zhao2023verify}).}
    \label{tab:additional_experiments}
\end{table}\\
% \vspace{-1pt}
~~~\textbf{Advanced Prompting and RAG} We conducted additional experiments on multi-hop PAT-Quesions using advanced prompting and RAG techniques. Table \ref{tab:additional_experiments} shows the experimental results on Chain-of-Thought (CoT) prompting (\citet{wei2022chain}), ReAct (\citet{yao2022react}), and Verify-and-edit \cite{ zhao2023verify}. We observe that CoT results in up to a 6.5\% increase in accuracy. This improvement is seen in questions whose answers have remained the same over the last year and for which the LLMs already have the most up-to-date knowledge. However, the overall performance is still very low, clearly indicating that the PAT-Questions dataset poses a significant challenge even for sophisticated prompting methods.

Unlike CoT, ReAct and Verify-and-edit perform worse for some questions, even with external knowledge access, due to temporally misaligned document retrieval. Although the RAG performance improves w.r.t. our GCS retrieval (7.6\%), the score is still lower than that of the direct prompting scores for GPT-3.5 (9.3\%). These methods show only 35-39\% accuracy in multi-hop QA tasks in general, highlighting the gap in multi-hop reasoning. We believe the lower performance with ReAct and Verify-and-edit is due to the additional complexity of present-anchoring. We also observe that these methods often generate reasonable intermediate steps i.e. one-hop questions but the retrieved facts are incorrect, which diverts the next steps from the ideal path (especially for `before'/`previous' temporal relations). %Our results highlight the performance gap of the advanced RAG methods in present-anchored multi-hop reasoning.
\begin{table}[t]
\small
    \centering
    \begin{tabular}{c|c|c}
    \toprule
     & single-hop & multi-hop\\
          \midrule
         Recall@5 & 19.24\%	& 15.40\% \\
         \midrule
         Recall@1 &	11.1\%	& 8.50\%\\
         \bottomrule
    \end{tabular}
    \caption{The evaluation of the retrieval quality of Google Custom Search (GCS)}
    \label{tab:retrieval_eval}
\end{table}\newline
\textbf{   GCS Retrieval Evaluation} Table \ref{tab:retrieval_eval} shows the Recall @top 5 and @top 1 for our Google Custom Search (GCS) retrieval. We observe that the recall scores are very low, especially @top 1, highlighting the gap in traditional retrievers in temporal reasoning, especially in PAT reasoning. 

\section{Conclusion}\vspace{-4pt}
In this paper, we introduced a novel self-updating dataset, PAT-Questions, of present-anchored temporal questions requiring both single and multi-hop reasoning on complex temporal relations. We provide a detailed evaluation in both direct prompting and RAG settings of the SOTA LLMs and TEMPREASON-T5 on PAT-Questions, and present the limitations of the LLMs in PATQA. The findings indicate a significant gap in LLMs' reasoning capabilities when addressing PAT-Questions. We provide an automatic answer updating system for the research community to retrieve the up-to-date answers of  PAT-Questions. 
% \newpage
\section*{Limitations}
Our self-updating system depends on an up-to-date knowledge base. We use the Wikidata knowledge base (KB), which may occasionally experience refreshing delays, potentially desynchronizing some gold annotations. 
Further, we retrieved documents for the PAT-Questions in our RAG pipeline solely using Google Custom Search API. However, this aspect is less significant given that our primary focus is not improving retrieval accuracy.
% However, improving the retrieval accuracy is not the focus of our work. 
Additionally, the scope of our multi-hop questions is currently limited to 2-hops, which already pose significant challenges for LLMs. We leave $2^{+}$-hop questions for future work.
\section*{Ethics Statement}
 We built our dataset entirely from publicly available information on Wikidata. No personal or restricted data were collected from any source or subject. Although the LLMs may sometimes generate fake information i.e. hallucinate, our experiments do not involve LLMs in creating any harmful content and, thus raise no ethical concern. We adhere to the Code of Ethics with our work. 
% \section*{Ethics Statement}

\section*{Acknowledgements}
This work was partially supported by NSF grants IIS-2227669 and IIS-1901379.

\bibliography{custom}

\appendix

\section{Templates details} \label{app:template_details}
We show our single-hop and multi-hop templates in Table \ref{tab:simpletemplates}
and \ref{tab:complextemplates} respectively. The Wikidata relation distribution over PAT-Questions is shown in Figure \ref{fig:rel_distribution}.
\begin{table*}[t]
\small
    \centering
    \begin{tabular}{p{0.23\textwidth}|p{0.54\textwidth}}
        \toprule
        \textbf{KB relation (Wikidata ID)}  & \textbf{Single-hop PAT-Questions Templates} \\
        \midrule
        \multirow{2}{*}{member of sports team (P54)}  & Which team does \{s\} play for currently?\\
        &  Which team did \{s\} play for before the current team?\\
        \midrule
        \multirow{2}{*}{position held (P39)} & Which position does \{s\} hold currently? \\
        &  Which position did \{s\} hold before the current position?\\
         \midrule
         \multirow{2}{*}{employer (P108)} & Which employer does \{s\} work for currently?  \\
         & Which employer did  \{s\} work for before the current employer?\\
         \midrule
         \multirow{2}{*}{political party (P102)} & Which political party does \{s\}  belong to currently? \\
         &  Which political party did \{s\} belong to before the current political party?\\
         \hline
         \multirow{2}{*}{head coach (P286)} & Who is the head coach of the team \{s\} currently?  \\
         & Who was the previous head coach of the team \{s\}? \\
         \midrule
         \multirow{2}{*}{chairperson (P488)} & Who is the chair of \{s\} currently?\\         
         & Who was the previous chair of \{s\}? \\
         \midrule
         \multirow{2}{*}{head of government (P6)} & Who is the head of the government of \{s\} currently?\\
         & Who was the previous head of the government of \{s\}? \\
         \midrule
         \multirow{2}{*}{head of state (P35)} & Who is the head of the state of \{s\} currently? \\       
         & Who was the previous head of the state of \{s\}? \\
         \midrule
         \multirow{2}{*}{owner (P127)} & Who is the owner of \{s\} currently? \\
         & Who was the previous owner of \{s\}? \\
         \bottomrule
    \end{tabular}
    \caption{Single-hop PAT-Questions Templates}
    \label{tab:simpletemplates}
\end{table*}

\begin{table*}[t]
\small
    \centering
    \begin{tabular}{p{0.28\textwidth}|p{0.5\textwidth}}
        \toprule
        \textbf{KB relation (Wikidata ID)} & \textbf{Multi-hop PAT-Questions Templates} \\
        \midrule
        \multirow{2}{*}{head coach (P286) } & Who is the head coach of the team that \{s\} plays for currently?\\
        & Who is the head coach of the team that \{s\} played for before the current team?\\
        \midrule
        \multirow{2}{*}{home venue (P115)} & What is the home venue of the team that \{s\} plays for currently?\\
        &  What is the home venue of the team that \{s\} played for before the current team?\\
        \midrule
        \multirow{2}{*}{owned by (P127)} & Who is the owner of the team that {subject} play for currently?\\
        &  Who is the owner of the team that \{s\} played for before the current team?\\
        \midrule
        \multirow{2}{*}{replaces (P1365)} & Who was the last person to hold the position that \{s\} holds currently?\\
        & Who was the last person to hold the position that \{s\} held previously?\\
        \midrule
        \multirow{2}{*}{spouse (P26)} & Who is the spouse of the current chair/owner/head of the state/head of the government of \{s\}?\\
        & Who is the spouse of the previous chair/owner/head of the state/head of the government of \{s\}?\\
        \midrule
        \multirow{2}{*}{educated at (P69)} & Which school did the current head coach/chair/owner/head of the government/head of the state of \{s\} attended?\\
        &  Which school did the previous head coach/chair/owner/head of the government/head of the state of \{s\} attended?\\
        \midrule
        \multirow{2}{*}{headquarters (P159)} & Where is the headquarters of the team that \{s\} plays for currently?\\
        & Where is the headquarters of the political party \{s\} belongs to currently?\\
        & Where is the headquarters of the current/previous owner of \{s\} located at?\\
        & Where is the headquarters of the current/previous employer of \{s\}?\\
        &  Where is the headquarters of the team that \{s\} played for before the current team?\\
        & Where is the headquarters of the political party that \{s\} belonged to before the current political party?\\
        \midrule
        \multirow{2}{*}{chair (P488)} & Who is the chair of the current employer of \{s\}?\\
        & Who is the chair of the political party which the \{s\} belongs to currently?\\
        &  Who is the chair of the previous employer of \{s\}?\\
        & Who is the chair of the political party which the \{s\} belonged to before the current political party?\\
        \midrule
        \multirow{2}{*}{chief executive officer (P169)} & Who is the chief executive officer  of the current employer of \{s\}?\\
        &  Who is the chair of the previous chief executive officer of \{s\}?\\
        \midrule
        \multirow{2}{*}{country of citizenship (P27)} & What is the country of citizenship of the current chair/head coach of \{s\}?\\
        &  What is the country of citizenship of the previous chair/head coach of \{s\}?\\
        \midrule
        \multirow{2}{*}{country (P17)} & Which country is the current owner of \{s\} from?\\
        & Which country is the previous owner of \{s\} from?\\
        \midrule
        \multirow{2}{*}{birthplace (P19)} & Where was the current head coach/head of the government/chair/owner of \{s\} born?\\
        &  Where was the previous head coach/head of the government/chair/owner of \{s\} born?\\
         \bottomrule
    \end{tabular}
    \caption{Multi-hop PAT-Questions Templates}
    \label{tab:complextemplates}
\end{table*}

\section{Dataset Statistics} \label{app:data_stat}
\begin{figure*}[t]
    \begin{subfigure}[b]{0.5\textwidth}
      \centering
      \includegraphics[width=\linewidth]{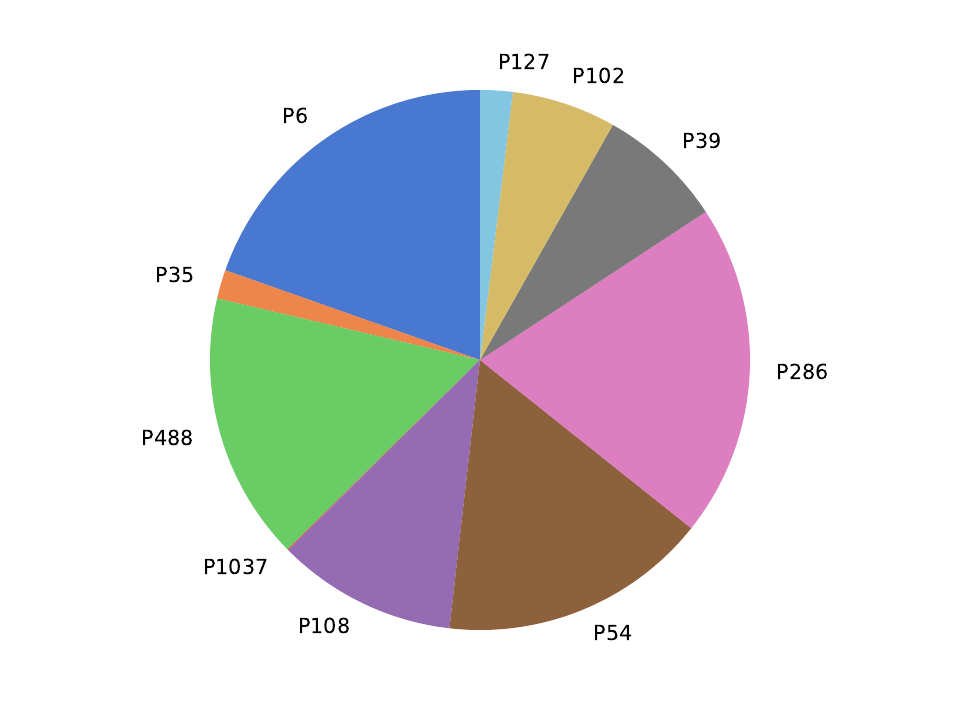}
      \caption{Relation distribution over single-hop PAT-Questions}
      \label{fig:nq_graph}
    \end{subfigure}%
    ~
    \begin{subfigure}[b]{0.5\textwidth}
      \centering
      \includegraphics[width=\linewidth]{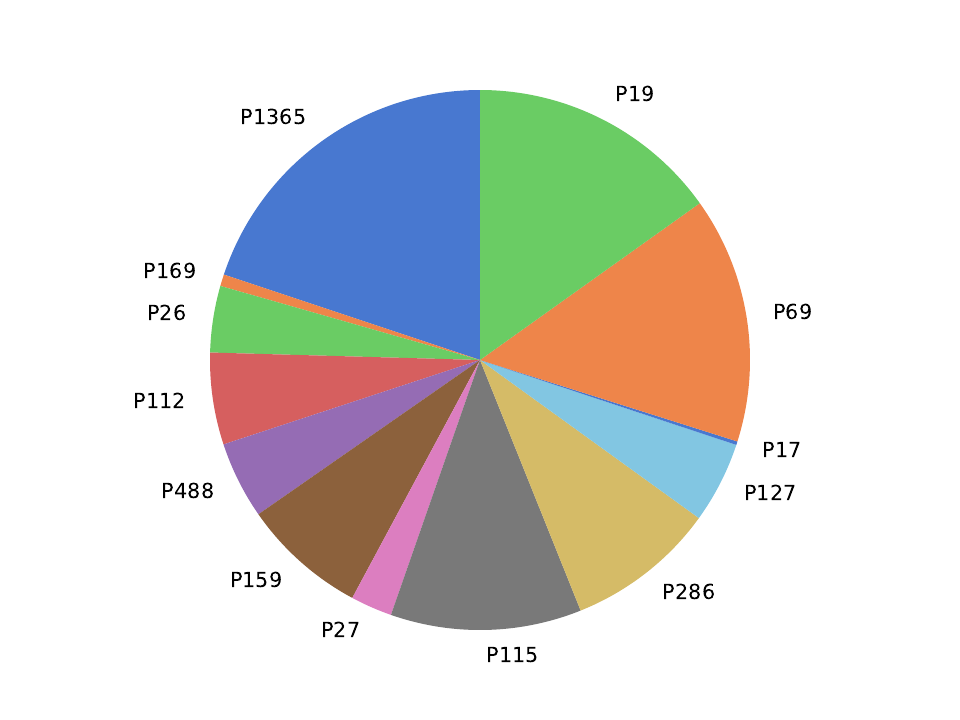}
      \caption{Relation distribution over multi-hop PAT-Questions}
      \label{fig:wq_graph}
    \end{subfigure}%
\caption{Fig. (a) and (b) show the Wikidata relation distributions over PAT-Questions}
    \label{fig:rel_distribution}
\end{figure*}

PAT-Questions contains 2882 single-hop and 3290 multi-hop questions with varied subjects and time-dependent relations from Wikidata. Each question in our dataset has seven common fields associated with it, `question', `subject', `text answers', `answer annotations', `relations', `template', and `uniq\_id'. Multi-hop questions have an extra field named `intermediate entities' denoting the one-hop answers to the multi-hop questions. An example is shown in Table \ref{tab:pat_example}.
\begin{table*}[t]
    \centering
    \begin{tabular}{p{0.8\textwidth}}
        \toprule
          `question': ``Who is the spouse of the current head of the state of Indonesia?'', \\
            ``subject'': \{``subject'': "Q252", ``subjLabel": "Indonesia"\},\\
            "text answers": [``Iriana" ],\\
            "answer annotations": [\{``ID": "Q17410605", ``Label": "Iriana"\}],\\
            "intermediate entities": [\{``ID": "Q3318231", ``Label": "Joko Widodo" \}],\\
            ``relations": [``P35", ``P26"],\\
            ``template": ``Who is the spouse of the current head of the state of \{subject\}?'',\\
            ``uniq\_id": 2986\\
        \bottomrule
    \end{tabular}
    \caption{An example multi-hop question from our PAT-Question dataset}
    \label{tab:pat_example}
\end{table*}
Figure \ref{fig:rel_distribution} shows the distribution of different Wikidata relations/properties over our dataset.
\section{SPARQL Templates} \label{app:sparql}
\begin{figure*}
    \centering
    \includegraphics[width=\linewidth]{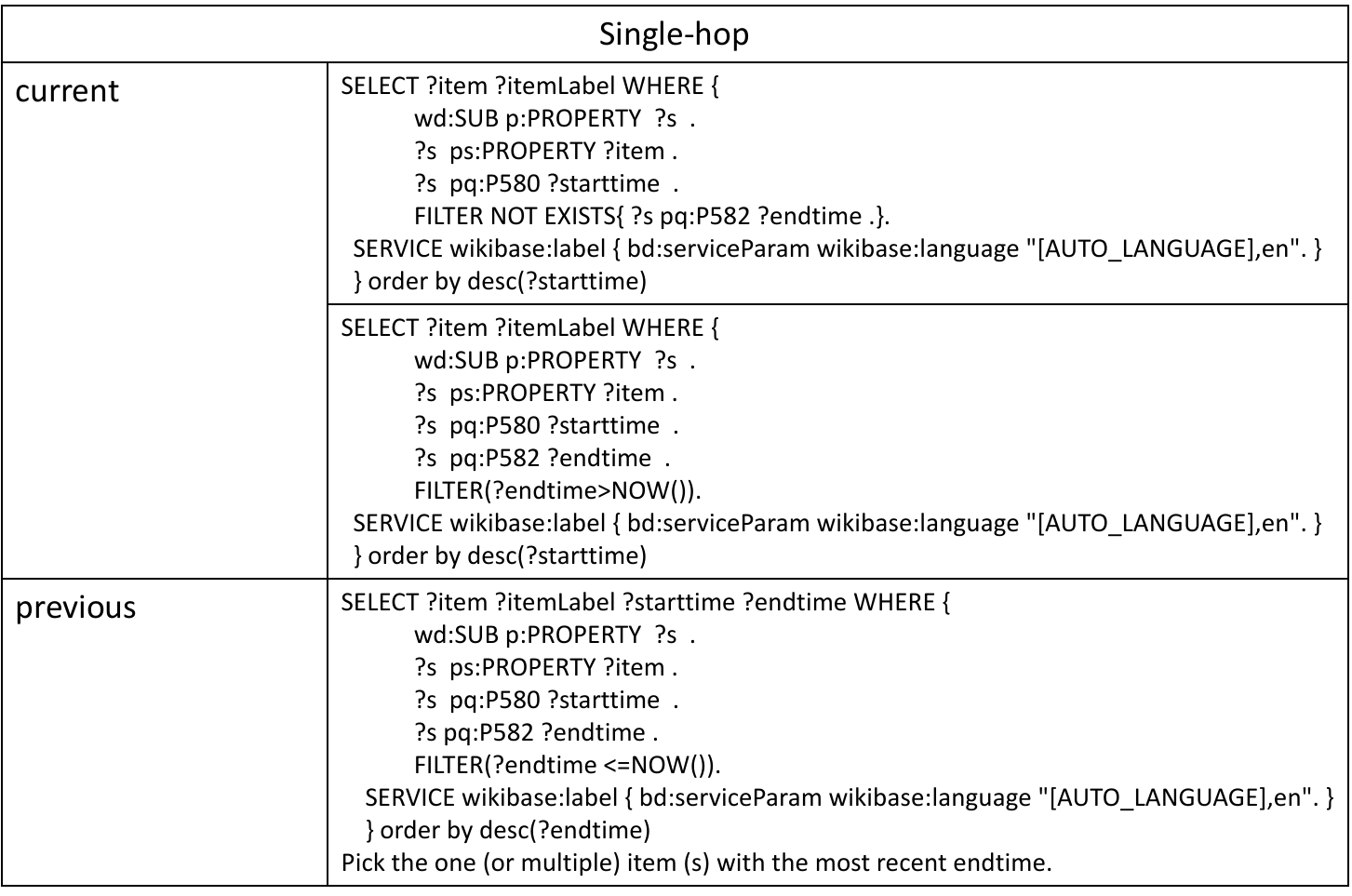}
    \caption{SPARQL templates for single-hop PAT-Questions' answer-update}
    \label{fig:single-sparql}
\end{figure*}

\begin{figure*}[t]
    \centering
    \includegraphics[width=\linewidth]{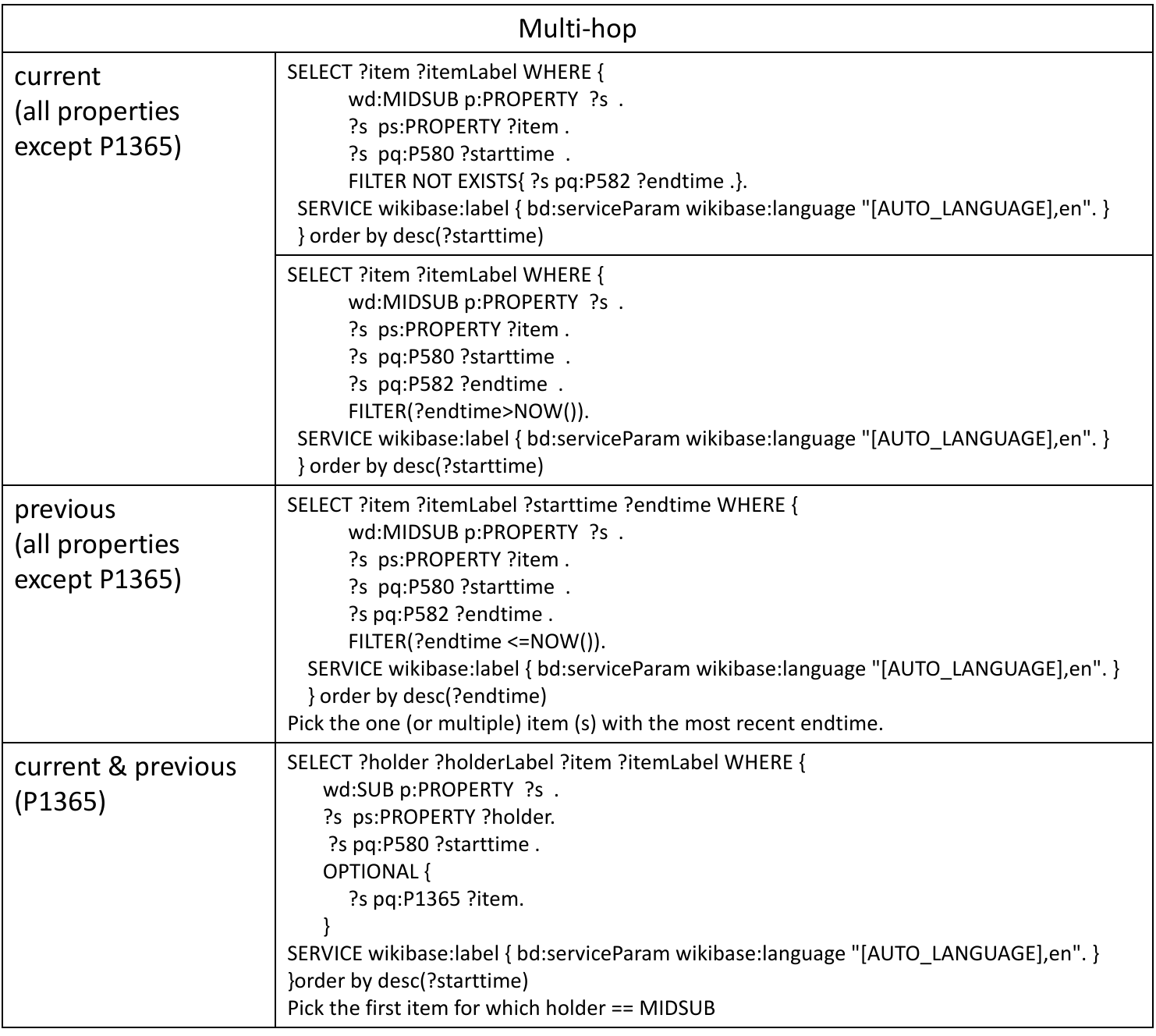}
    \caption{SPARQL templates for multi-hop PAT-Questions' answer-update}
    \label{fig:multi-sparql}
\end{figure*}

We update the answers to our dataset via running SPARQL queries over Wikidata, and retrieve the most up-to-date answer with respect to knowledge available/updated in Wikidata. Our SPARQL templates are illustrated in Figure \ref{fig:single-sparql} and \ref{fig:multi-sparql}. Some cases cannot be handles directly via the SPARQL queries. For example, for some facts `P1365' qualifier does not exist. For such cases, we have a special clause in our script for extracting the up-to-date answer.

\section{Sample Prompts and Responses of LLMs} \label{app:prompts}
\begin{figure*}
    \centering
    \includegraphics[width=\linewidth]{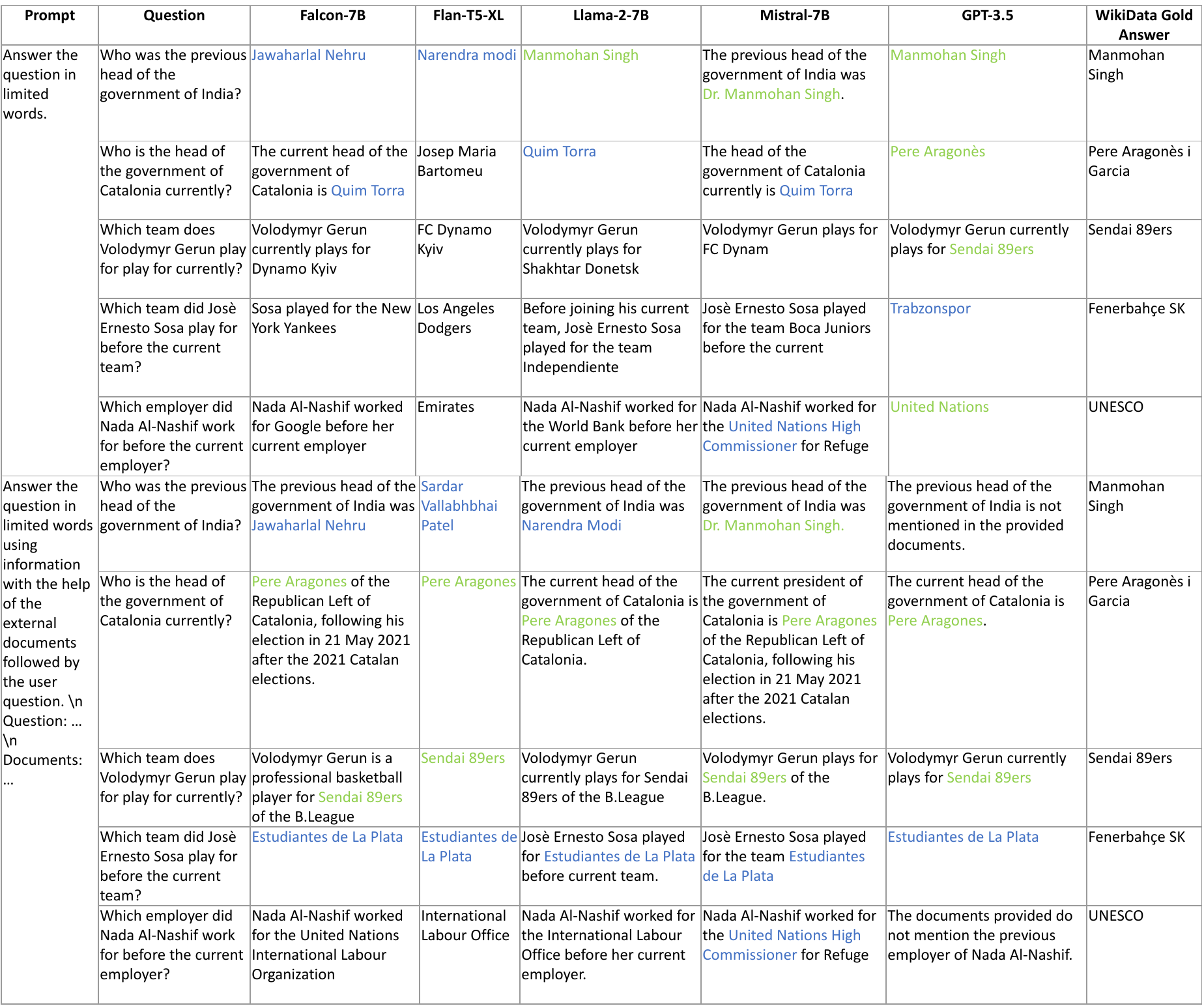}
    \caption{Responses of LLMs to single-hop PAT-Questions when asked without and with RAG.}
    \label{fig:singlehop_example}
\end{figure*}

\begin{figure*}
    \centering
    \includegraphics[width=\linewidth]{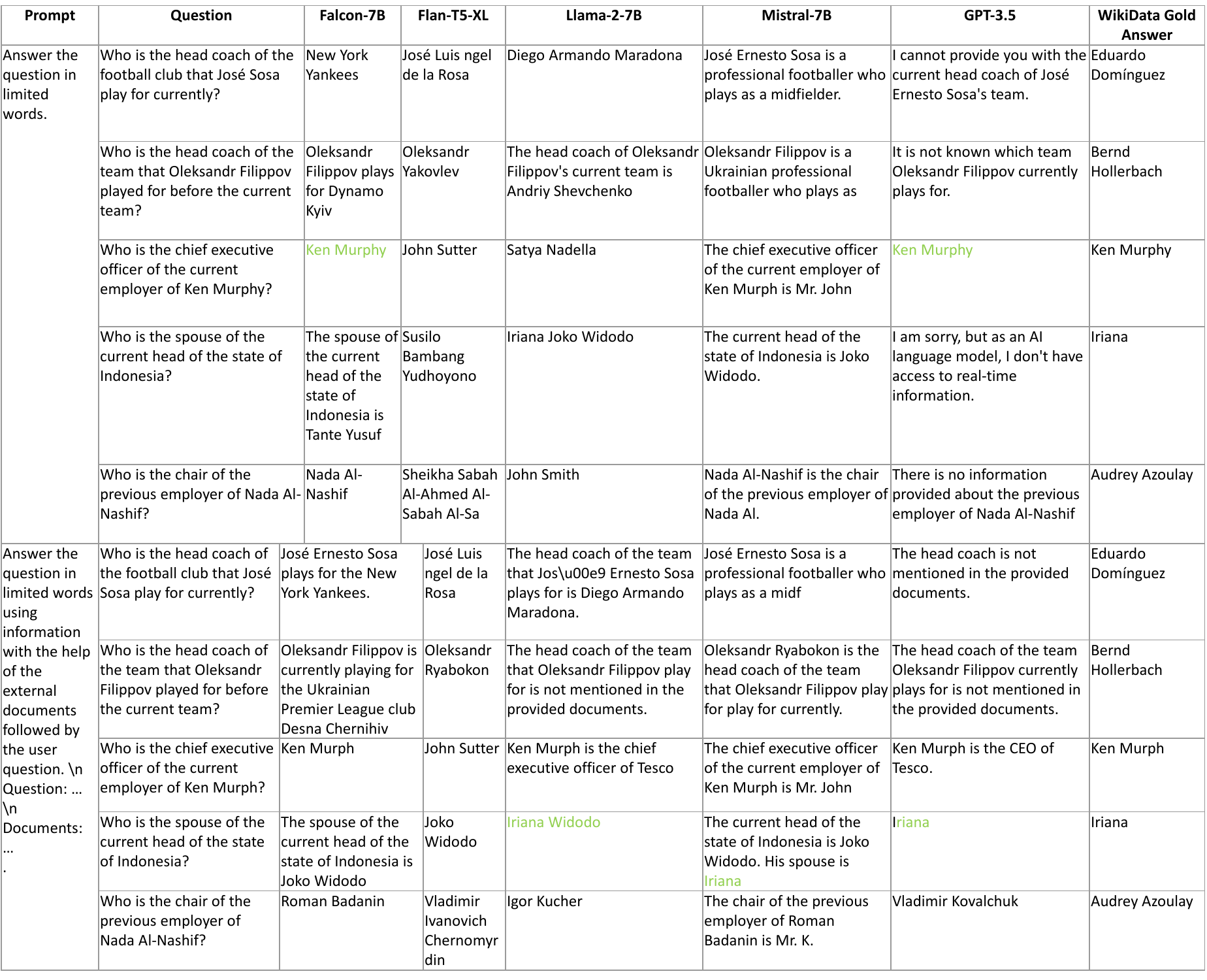}
    \caption{Responses of LLMs to multi-hop PAT-Questions when asked without and with RAG.}
    \label{fig:multihop_example}
\end{figure*}

\begin{figure*}
    \centering
    \includegraphics[width=\linewidth]{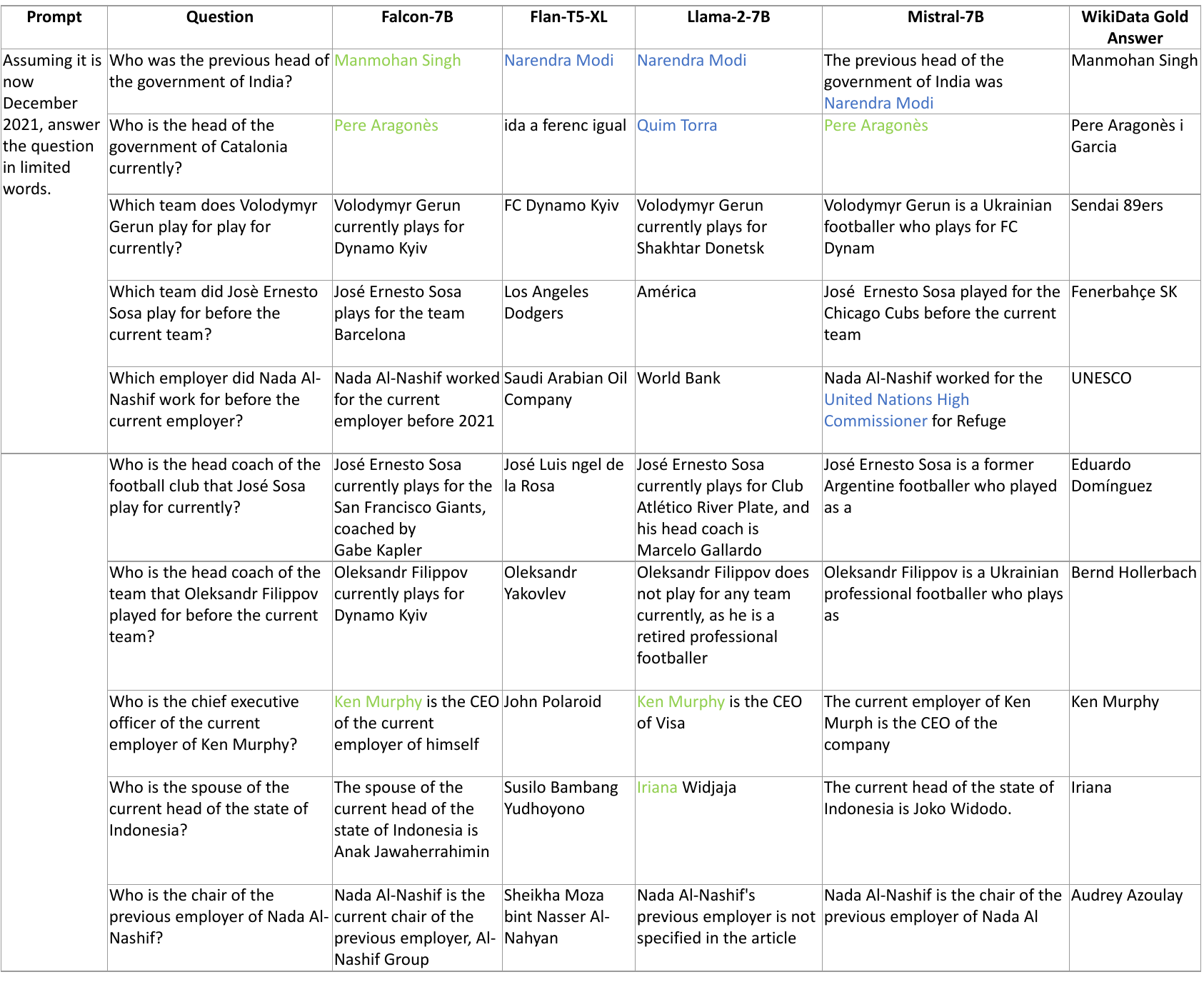}
    \caption{Responses of LLMs when prompted with questions prepending ``Assume it is not December 2021,'' to the prompt.}
    \label{fig:assume21_example}
\end{figure*}

We show the prompt for our experimental settings and some sample responses of the LLMs in Figures \ref{fig:singlehop_example}, \ref{fig:multihop_example} and \ref{fig:assume21_example}. Responses marked in green are the correct answers, blue are the correct answers from a different point in time, and orange are the correct answers to the one-hop questions leading to incorrect or no response to the multi-hop questions. Open-source LLMs tend to provide more relevant and accurate answers compared to when directly prompting the question. This extra instruction helps them ground their answers on the provided time. LLMs' performance improves when one or more relevant passages are provided as context in the RAG setting. As shown in Figures \ref{fig:singlehop_example} and \ref{fig:multihop_example}. However, if the retrieved passages are irrelevant, the accuracy may drop, such as for the question ``Who was the previous head of the government of India?'' in Figure \ref{fig:singlehop_example}, GPT-3.5 fails to answer the question in the RAG setting whereas responds correctly to the same question when asked directly.

Although utilizing LLMs can be quite expensive, as closed source LLMs charge based on the number of tokens \cite{rashid2024ecorank}, for our experiments, we tried different prompts such as directly asking the question without any instructions, instructing the LLMs to ``Answer the question'', ``Answer the question in limited words'', ``Answer the question in a single phrase'' etc. Out of the prompts, ``Answer the question in limited words'' resulted in the best results when compared to the EM accuracy. 

\subsection*{Sample Responses of New Bing and GPT-4 Web}
\begin{figure*}
    \centering
    \includegraphics[width=\linewidth]{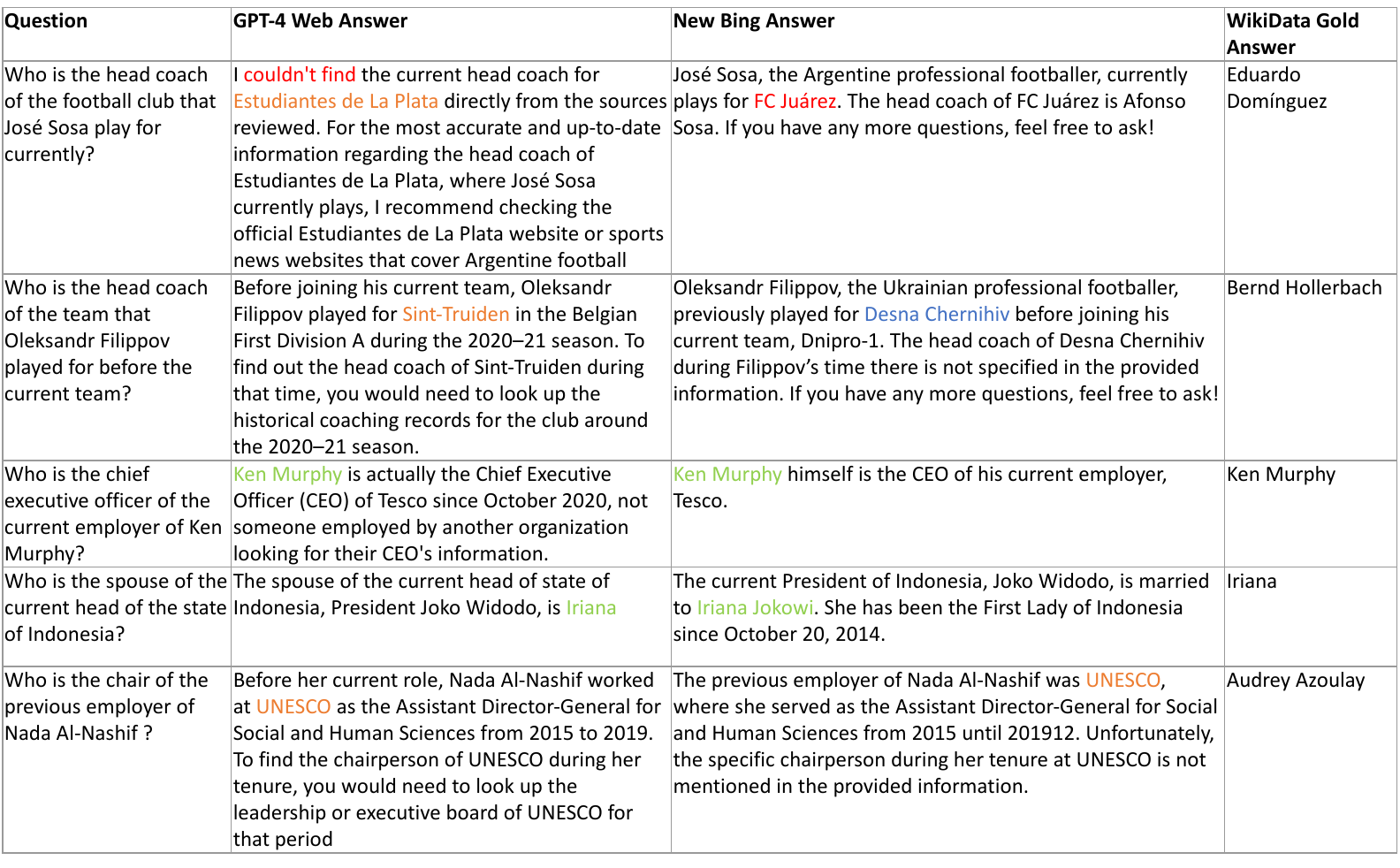}
    \caption{BingChat and GPT responses along with the gold answer for 5 sample multi-hop questions of our dataset. Correct answers are marked in green, outdated answers are marked in blue, correct one-hop answers but unable to find the two-hop answers are marked in orange and completely false or irrelevant answers are marked in red.}
    \label{fig:bing-gpt}
\end{figure*}

We show some sample responses of BingChat and GPT-4 Web to our multi-hop PAT-Questions in Figure \ref{fig:bing-gpt}. Although these services have access to the current knowledge, they sometimes provide outdated answers (blue), correct one-hop answers but cannot find the two-hop answers (orange), completely false answers (red), or fail to understand the questions and respond with some other facts about the subjects (red).
\begin{comment}
\begin{table*}[t]
    \centering
    \begin{tabular}{|c|c|c|c|}
    \hline
         \multirow{2}{*}{KB relation} & \multirow{2}{*}{Type} & \multicolumn{2}{c|}{Category} \\
         \cline{3-4}
         & & Simple & Complex\\
         \hline
         \multirow{2}{*}{member of sports team} & current & x & y\\
         \cline{2-4}
         & before-current & x & y\\
         \hline
         \multirow{2}{*}{employer} & current & x & y\\
         \cline{2-4}
         & before-current & x & y\\
         \hline
         \multirow{2}{*}{political party} & current & x & y\\
         \cline{2-4}
         & before-current & x & y\\
         \hline
         \multirow{2}{*}{head coach} & current & x & y\\
         \cline{2-4}
         & before-current & x & y\\
         \hline
         \multirow{2}{*}{chairperson} & current & x & y\\
         \cline{2-4}
         & before-current & x & y\\
         \hline
         \multirow{2}{*}{head of government} & current & x & y\\
         \cline{2-4}
         & before-current & x & y\\
         \hline
         \multirow{2}{*}{head of state} & current & x & y\\
         \cline{2-4}
         & before-current & x & y\\
         \hline
         \multicolumn{2}{|c|}{Total \# question} & 3170 & 2882\\
         \cline{3-4}
         \multicolumn{2}{|c|}{} & \multicolumn{2}{c|}{5633}\\
         \hline
    \end{tabular}
    \caption{Dataset Statistics}
    \label{tab:data_stat}
\end{table*}
\end{comment}

\section{Filter for incorrect responses} \label{filter_error}
For our error analysis, we extracted all objects associated with our time-dependent facts along with their aliases to check if an LLM response matches any of the outdated results. To filter responses like `information not available' or `cannot respond', we manually went through a subset of LLM responses and picked some keywords that exist in such responses such as `cannot provide information', `unknown', `N/A' etc. Any response that does not fall into the set of extracted outdated objects or `cannot respond' filter are considered as false or fake, or incomplete response.

\label{sec:appendix}

\end{document}